# Development of a deep learning platform for optimising sheet stamping geometries subject to manufacturing constraints

Hamid Reza Attar[a], Alistair Foster[b], Nan Li[a,*]

[a] *Dyson School of Design Engineering, Imperial College London, London SW7 2DB, UK*
[b] *Impression Technologies Ltd, Coventry CV5 9PF, UK*

*Corresponding author
E-mail address: n.li09@imperial.ac.uk (N.Li)

## Abstract

The latest sheet stamping processes enable efficient manufacturing of complex shape structural components that have high stiffness to weight ratios, but these processes can introduce defects. Considering stamping capabilities during component design phases can reduce such defects beyond what is achievable through tuning process settings alone. To assist component design for stamping processes, this paper presents a novel deep-learning-based platform for optimising 3D component geometries. The platform adopts a non-parametric modelling approach that is capable of optimising arbitrary geometries from multiple geometric parameterisation schema. This approach features the interaction of two neural networks: 1) a geometry generator and 2) a manufacturing performance evaluator. The generator predicts continuous 3D signed distance fields (SDFs) for geometries of different classes, and each SDF is conditioned on a latent vector. The zero-level-set of each SDF implicitly represents a generated geometry. Novel training strategies for the generator are introduced and include a new loss function which is tailored for sheet stamping applications. These strategies enable the differentiable generation of high quality, large scale component geometries with tight local features for the first time. The evaluator maps a 2D projection of these generated geometries to their post-stamping physical (e.g., strain) distributions. Manufacturing constraints are imposed based on these distributions and are used to formulate a novel objective function for optimisation. A new gradient-based optimisation technique is employed to iteratively update the latent vectors, and therefore geometries, to minimise this objective function and thus meet the manufacturing constraints. Case studies based on optimising box geometries subject to a sheet thinning constraint for a hot stamping process are presented and discussed. The results show that expressive geometric changes are achievable, and that these changes are driven by stamping performance. The platform has the potential to guide necessary geometric design changes at the beginning of a design process and thus encourage the adoption of the latest sheet stamping processes.

## 1 Introduction

### 1.1 Industrial context

Every year, thousands of structural components are designed and manufactured through sheet stamping processes. Sheet stamping processes offer several advantages over other manufacturing processes such as machining or casting [1]. These advantages include low production cost at scale, high production efficiency and high stiffness to weight ratios.

The high stiffness to weight ratio components that are achievable through these processes satisfy the rising demand for lightweighting in the automotive and other transport industries [2]. Aluminium alloys are a family of lightweight metals that offer exceptional specific strength and stiffness and therefore the stamping of these alloys is promising for further meeting this lightweighting demand. Despite several advantages over conventional heavier steels for sheet metal stamping applications, aluminium alloys suffer from poor formability at room temperature and springback after the removal of forming tools [3,4].

These issues could be significantly reduced by adopting the state-of-the-art Hot Forming and cold die Quenching (HFQ®) stamping process developed by Lin *et al.* [5]. During this process, aluminium alloy blanks are heated to their solution heat treatment (SHT) temperature before being simultaneously stamped and quenched in cold dies. Further details on the HFQ process can be found in [4,5].

However, although the designed elevated temperature conditions improve the material formability, there are several inherent manufacturing complexities associated with the HFQ process. The temperature difference due to the cold forming tools and heated blank makes the process non-isothermal which results in uneven cooling during stamping [6]. The flow stress response of the material at elevated temperatures is highly strain rate dependent [4,7]. In addition, the temperature and strain rate vary dynamically with time and location in the sheet metal which is unlike conventional cold stamping processes [8]. These complexities not only create an unfamiliarity among industrial component designers but also limit the space of feasible component designs. Consequently, the uptake of HFQ and similar hot stamping processes [9] is limited in modern industrial settings. As a result, the advantages of these sheet stamping processes are currently not realised to their full potential.

## 1.2 Computer methods for manufacturability assessment of stamping geometries

Previous research on HFQ has largely focused on the development of advanced material models that are capable of predicting the material constitutive behaviour under HFQ conditions [4,7,10–12]. The intention is to embed these models into Finite Element (FE) numerical simulations to accurately simulate the stamping behaviour of components through HFQ [6,11,13]. However, these simulations usually take place late in design processes and often when component design is near completion [14,15]. In addition, the computationally costly simulator would have to be run by a process engineer with extensive stamping process expertise each time a design engineer wishes to make a geometry change. This continuous rerun and ongoing consultation between process and design engineers makes the simulation route slow and costly and therefore practically unsuitable for early stage design optimisation.

To provide a more efficient optimisation procedure, numerical optimisation algorithms are available. Existing literature on numerical optimisation for sheet stamping largely focuses on optimisation of process parameters for a given fixed geometry. For example, Xiao *et al.* [16] used a stochastic approach to optimise stamping speed, blank holder force, friction and blank temperature when forming an automotive floor component under HFQ conditions. Similarly, Zhou *et al.* [17] use a genetic algorithm to optimise blank holder force and stamping speed to minimise thinning and springback defects in a anti-collision side beam component under HFQ conditions. This heavy focus on process parameters is because stamping process capabilities are considered late in design processes where geometries have already been designed, as explained by Attar *et al.* [15]. At this stage, the design is passed to process engineers, who are tasked with modifying process parameters to minimise stamping induced defects for the given geometry. In contrast, considering stamping capabilities during early component design phases can reduce such defects beyond what is achievable through tuning process settings alone [14].

The arrival of machine learning (ML) has introduced a new way to establish time efficient process models that can convey manufacturability information to early geometry design phases. These models are therefore a promising route towards tackling the issue of stamping process unfamiliarity and can provide new insights into stamping behaviour. In this context, researchers have recently developed ML models that serve as surrogate models of stamping processes. Surrogate models typically learn system dynamics from sourced legacy data and are then employed to make predictions on unseen scenarios [18]. Dib *et al.* [19] used several surrogate models to predict the springback and maximum thinning of a U-channel and square cup stamping process. Harsch *et al.* [20] used surrogate models to develop processing parameter window maps for a cold stamping process. Liu *et al.* [21] used a novel theory-guided neural network surrogate model to predict displacement after a cold sheet metal bending process. However, those models rely on hand-crafted parameterisations as inputs and are only effective in scenarios where the design space consists of relatively few parameters. Examples of such scenarios include development of design guidelines for simplified approximations of stamping geometries [15] and the optimisation of processing parameters for fixed geometries [17].

Unlike processing parameters, geometries are significantly more varied, frequently evolving during early stage design and defined according to uncommon CAD parameterisation schema. These challenges have recently brought to light surrogate models that consider geometric inputs in a non-parametric way by borrowing modelling techniques from the field of deep learning. In particular, convolutional neural networks (CNNs) have emerged as a powerful tool for predicting physics-based performance of geometry-based data. CNNs are a particular class of neural network that are particularly efficient when working with spatially structured data such as grids of images. Hao *et al.* [22] demonstrated the superior accuracy achieved by CNN based surrogate models over more conventional Kriging methods when predicting the buckling performance of curvilinearly stiffened panels. Researchers have also found CNNs to be effective non-parametric surrogate models in several other domains, including solid mechanics [23,24], wave physics [25,26] and computational fluid dynamics [27,28].

In the domain of sheet stamping, Zimmerling *et al.* [18,29,30] to use several different image-based surrogate models to predict the textile draping results when varying geometries. For the HFQ sheet metal stamping processes, Attar *et al.* [31] developed a novel CNN-based surrogate model that considered variations in stamping die geometry. In their work, a CAD geometry was projected onto an image from which the model was able to predict the post-stamping physical fields associated with the geometry inexpensively and with high accuracy. Furthermore, Attar *et al.* [31] found that this high accuracy was achievable even though the underlying HFQ stamping process had the aforementioned non-uniform temperature and strain rate complexities. CNN-based surrogate models are therefore a promising route for optimising component geometries for stamping processes during early design phases and thus maximising stamping process capabilities.

However, CNN-based surrogate models alone are only capable of forward predictions for variable geometries. This limitation means that designers would need to implement a trial-and-error approach since there is no backward feedback or guidance towards optimum geometric designs. Moreover, since these models take image representations of geometries as inputs, there are no parameters available to serve as design variables for optimisation of stamping geometries. This drawback currently prevents CNN-based surrogate models from being used in numerical optimisation settings. Therefore, there is a need for a computer based representation of stamping geometries which when combined with CNN-based surrogate models enables stamping performance driven optimisation of arbitrary geometries.

## 1.3 Computer methods for geometric representation

Modern image-based representation learning techniques are available for generating realistic images from compact optimisation-friendly representations. A popular approach is to use generative models such as variational auto-encoders (VAEs) [32,33] and generative adversarial networks (GANs) and their variants [34–36]. VAEs are trained to replicate variants of the original input but are prone to blurred reconstructed images. On the other hand, GANs learn deep embeddings of target data by training image generating decoders adversarially against discriminators. Once trained, these networks can generate realistic images of objects and scenes which are visually indistinguishable from their training data distributions [37]. However, the training of GANs is notoriously unstable [38,39] and difficult to extend to 3D geometric representations [40], e.g., high resolution 3D CAD geometries for designers. These major limitations make GANs unsuitable for practical stamping applications in industrial settings.

Several computer methods for geometric representation exist in the 3D geometric modelling community and these can generally be categorised as being either *explicit* or *implicit*. *Explicit* representations directly parametrise the geometry and are widely used in problem-specific engineering applications. In the field of CAD modelling, parametric spline-based curves are a commonly used explicit representation in modern CAD packages, e.g., SolidWorks. These curves are modelled as explicit functions and the parameters of these functions can be manipulated in an optimisation setting. For example, Li *et al.* [41] constructed and optimised addendum surfaces for sheet metal stamping using parametric curves. However, since explicit functions define these curves, this representation is significantly restricted to pre-defined topologies. Further, these functions lack expressivity, may be difficult for formulate for complicated geometries and do not allow for optimisation between different parameterisation schema.

A promising group of commonly used explicit representations are mesh-based representations, which store geometric information as a list of vertices and connected faces. By using fine mesh resolutions, high frequency geometric details on single shapes can be well represented, as seen from the examples provided by Sorkine *et al.* [42]. In attempts to extend these mesh-based representations to multiple arbitrary shapes, various works proposed representing meshes using data-driven 3D learning approaches, such as neural networks [43–45]. These networks represent classes of similar shapes by optimising/deforming initial template meshes. In this context, Wang *et al.* [45] proposed a network that generates a mesh from an image of an object by deforming the vertices of an initial spherical template mesh. Similarly, Baque *et al.* [46] deformed mesh vertices via a poly-cube mapping algorithm for shape optimisation. However, although meshes are capable of representing high frequency details, the results of mesh-based optimisations lose these details and often result in poorly deformed meshes. This phenomenon arises because the optimisation is based on deforming an initial predefined topology and therefore cannot handle large topology changes well, as mentioned by Peng *et al.* [47] and Park *et al.* [48].

In contrast, *implicit* representations define shapes as level sets of 3D functions over voxelised grids [47,49,50], and these level sets are then extracted or rendered into explicit meshes or images [49,51]. These representations easily allow large changes in topology as they remove the need to deform initial meshes. For example, Remelli *et al.* [52] changed the topology of genus 0 shapes into genus 1 by manipulating the underlying 3D function. However, operating on voxelised grids leads to large memory requirements and is therefore limited to low resolutions only. This limitation has been removed in recent years with the introduction of *implicit neural* representations which represent shapes as level sets of occupancy fields [53] or signed distance fields (SDFs) [48,54,55] that are generated from neural networks. These networks learn compact latent representations which are then decoded into their underlying 3D fields and these compact representations are suitable for optimisation.

Implicit neural representations have gained tremendous popularity in the computer graphics and vision communities for modelling 3D shapes because of their expressiveness and flexibility that is not limited by resolution [48]. These representations appear promising as they enable large topology changes in an optimisation setting [47]. However, the application focus of implicit neural representations is currently limited to scenarios where only the visual shape quality on a global scale is important. For example, modelling sofas and chairs in 3D scenes for computer games or virtual reality [48,56], or modelling 3D human poses [57,58]. In contrast to these geometries, implicit neural representations they have not been used for modelling nor optimising stamping geometries before. For stamping geometries, features such as fillet radii are highly influential in dictating manufacturing performance. For example, Attar *et al.* [15] showed that a small change in die and punch fillet radii produces a large change in sheet thinning during stamping. Despite their strong influence, fillet radii are small local geometric features when considered in the context of the global component scale. For example, automotive door inner components could be as wide as 1500 mm [13] and the latest stamping processes [4] can enable fillet radii of 10 mm or smaller. Therefore, representing stamping geometries for manufacturing performance evaluation and optimisation demands high quality reconstructions of small local features.

## 1.4 Summary and this contribution

In summary, the latest stamping processes that offer promising routes to vehicle lightweighting are unfamiliar among industrial designers which increases the component design development time. Comprehensive surrogate models have recently been developed to convey stamping process capabilities to early stage design phases but lack an optimisation-friendly geometric representation. Powerful implicit neural representations have recently emerged which allow for expressive topology changes in optimisation settings but have not been used for stamping applications before.

The work reported in this paper aims to address all three of these shortcomings. A comprehensive study on the development of a new deep-learning based platform for optimising sheet stamping geometries subject to manufacturing constraints is presented. The main contributions of this study are listed below:

- The development of a deep-learning methodology for generating high quality implicit neural representations of sheet stamping geometries with tight local features (e.g., fillet radii).

- The implementation of the newly developed methodology by training a neural network to reconstruct stamping geometries from multiple parameterisation schema. High reconstruction accuracy of local geometric features, which are critical to sheet stamping performance, were enabled.

- A comprehensive quantitative and qualitative evaluation to verify the geometric reconstruction accuracy.

- The extension of CNN-based surrogate models of hot stamping processes to geometries from multiple parameterisation schema. The developed model was able to predict post-stamping thinning fields from which manufacturing constraints were defined.

- The construction of a new optimisation platform from the interaction of the two aforementioned networks. The platform featured a novel backward pass which made it possible to use gradient based optimisation to update geometries in accordance with their manufacturing performance. Moreover, the combined use of implicit neural representations and CNN-based surrogate models meant that restrictions on geometry parameterisations were removed. This removal enabled the geometry to deform with complete freedom.

- The implementation of the platform to optimise deep drawn corner geometries subject to a maximum post-stamped thinning constraint. This implementation was able to non-parametrically update geometries between geometry subclasses, providing evidence that the optimisation platform is agnostic to geometric complexity.

## 2 Overview of the proposed design optimisation platform

An overview of the proposed platform for optimising component geometries subject to manufacturing constraints is introduced in this section. A high level explanation of the key aspects which make up the platform is given here, and specific details are explained in depth later in the paper. The setup configuration for the platform is then presented and explained.

### 2.1 Introduction to Signed Distance Fields (SDFs)

Signed Distance Fields (SDFs) are an important part of the non-parametric modelling strategy proposed in this paper, and hence a brief introduction of them is given here. SDFs are widely used in the computer vision and graphics communities to model a variety of shapes. Some examples of shapes modelled by SDFs include complex shaped 3D objects, and scenery in computer game environments. This work proposes using SDF to model stamping die surfaces.

#### 2.1.1 Definition of SDFs

A SDF for a shape with internal domain $\Omega$ is defined as a scalar field $s$ where the magnitude of a point anywhere in the field represents the distance $d$ to the shape boundary $\partial\Omega$. The distance $d$ is defined here as the Euclidian distance to the closest point sampled on the continuous shape boundary $\partial\Omega$. The sign denotes whether the point is inside or outside $\Omega$; by convention, points inside take a negative sign ($-$) and points outside take a positive sign ($+$) [48], as expressed in Equation (1)

$$s(x) = \begin{cases} 0, & x \in \partial\Omega \\ -d(x), & x \in \Omega \text{ and } x \notin \partial\Omega \\ d(x), & x \notin \Omega \end{cases} \quad (1)$$

where $s(x) \in \mathbb{R}$ is a scalar signed distance value of a point $x \in \mathbb{R}^n$ for an $n$-dimensional SDF. The shape boundary $\partial\Omega$ is implicitly represented by the zero-level-set of the continuous SDF. The classification space (($-$) or ($+$)) is explicitly represented by all other non-zero-level-sets.

Figure 1 shows 2D examples of SDFs for three different shapes: a circle, a square and an arbitrary shape. The zero-level-set of these SDFs are highlighted by the black solid lines, which would be surfaces in the 3D case. The figure illustrates that SDFs can be used to implicitly represent a range of shapes irrespective of their geometric complexity.

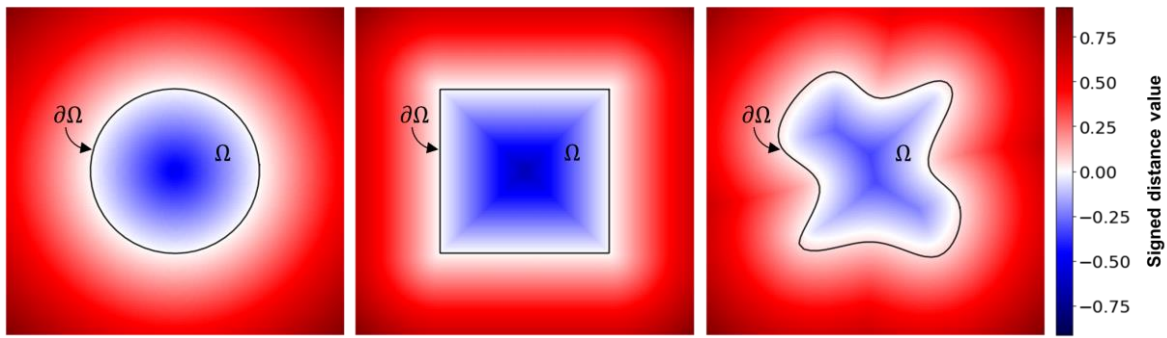

Figure 1 2D examples of SDFs for a circle, a square and an arbitrary shape. Black solid lines illustrate shape boundaries which are implicitly represented by the zero-level-set of the SDF.

#### 2.1.2 Geometric properties of SDFs

The shape representative nature of SDFs leads to two noteworthy geometric properties. **Property 1:** if the shape domain $\Omega$ is a subset of an $n$ dimensional Euclidian space $\mathbb{R}^n$ with shape boundary $\partial\Omega$, then the SDF is differentiable almost everywhere and its gradient satisfies the solution to the Eikonal partial differential equation, as expressed in Equation (2)

$$\|\nabla_x s(x)\| = 1 \quad (2)$$

where the $\nabla_x$ operator denotes the spatial gradient of $s(x)$ with respect to the coordinates of point $x$. To interpret this property, Figure 2 shows a 2D SDF for a circle shape but plotted with a discrete colourmap. The equidistant contour lines can be seen as a visualisation of constant spatial gradient. **Property 2:** the spatial gradients of the SDF at the shape boundary $\partial\Omega$ align with the outward normal vector field $\mathcal{N}$. This property arises since the gradient vector points in the direction of steepest descent (i.e., normal to each contour line). Furthermore, since Property 1 holds true everywhere (including at the shape boundary $\partial\Omega$), these surface normal vectors have unit magnitude. Property 2 can be mathematically expressed as in Equation (3).

$$\nabla_x s(x)|_{\partial\Omega} = \mathcal{N}(x) \tag{3}$$

Figure 2 shows the outward unit normal vectors at the shape boundary, where only a limited number are drawn for illustration. When combining Properties 1 and 2, the continuous SDF can be interpreted as a differentiable extension of the unit normal vector field. These properties will be used later in this paper to effectively train a neural network to learn a range of SDFs for a class of geometries and with high accuracy.

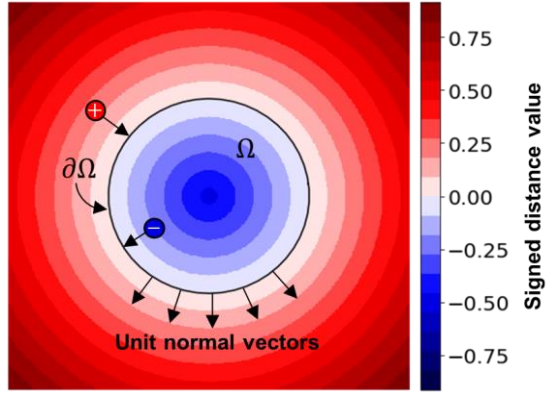

Figure 2 An illustration of a 2D SDF for a circle with labelled characteristic features for the SDF. Although the field is continuous, the colourmap is shown as discrete to highlight equidistant contour lines. Black solid line illustrates the shape boundary $\partial\Omega$ which is implicitly represented by the zero-level-set of the SDF. Points inside $\Omega$ have a negative distance to $\partial\Omega$ and points outside have a positive distance.

### 2.1.3    Modelling SDFs of stamping geometries

Recall that SDFs were used in this work to model stamping die surfaces. However analytically pre-computing and storing a large library of SDFs for various candidate geometries is neither feasible nor useful for optimisation. Instead, a model that can represent a wide range of geometries, discover similarities between them and store these similarities into a latent (i.e., characteristic) space is required.

To achieve this requirement, a neural network model was trained to generate full continuous SDFs of entire shape classes. This concept was first introduced by Park *et al.* [48] for generic shapes and these representations are commonly referred to in literature as *implicit neural representations*. Specific training details will be provided in depth later in Section 4. For stamping applications, the trained network can be considered as a compact function that generates SDFs of common classes of components, while only explicitly storing the parameters of the neural network. For example, a network can be trained to generate SDFs for thousands of variants of A-Pillars, B-Pillars, Battery Boxes, and similar broad classes of component geometries. In this work, box corner geometries were considered for demonstration.

To effectively work with CAD geometries of die surfaces in a deep learning environment, these geometries must first be mapped into a suitable form. Here, each CAD geometry was represented as a latent vector $z \in \mathbb{R}^L$ as illustrated in Figure 3(a). These vectors were obtained using a process introduced as *decoder inference*, which will be detailed in Section 4.2. In essence, these latent vectors became compact characteristic representations of CAD geometries and were agnostic to shape complexities.

An trained Auto-Decoder [48] network $f_{\theta_1}$ with network parameters $\theta_1$, was then used to approximate an SDF $s_i$ for some geometry indexed by $i$, given its latent vector (i.e., for its CAD geometry) $z_i$. This formulation is denoted in Equation (4) for a query point $x \in \mathbb{R}^3$, and is valid for all points within the considered SDF volume.

$$f_{\theta_1}(z_i, x) \approx s_i(x), \forall x \tag{4}$$

Figure 3(b) illustrates a generic Auto-Decoder. The (x, y, z) spatial coordinates of a query point in ambient 3D space $x$ were first concatenated with a given latent vector. This concatenated vector was decoded by the decoder into a scalar SDF value at the input 3D query point. Performing this forward pass on a full grid of points $X$ would generate the entire SDF $f_{\theta_1}(z_i, X)$ over that grid. This generated SDF would be conditioned on the latent vector for the geometry indexed by $i$. Therefore, changing the latent vector (i.e., in an optimisation setting) would change the generated SDF, and this phenomenon is exploited in the proposed optimisation platform.

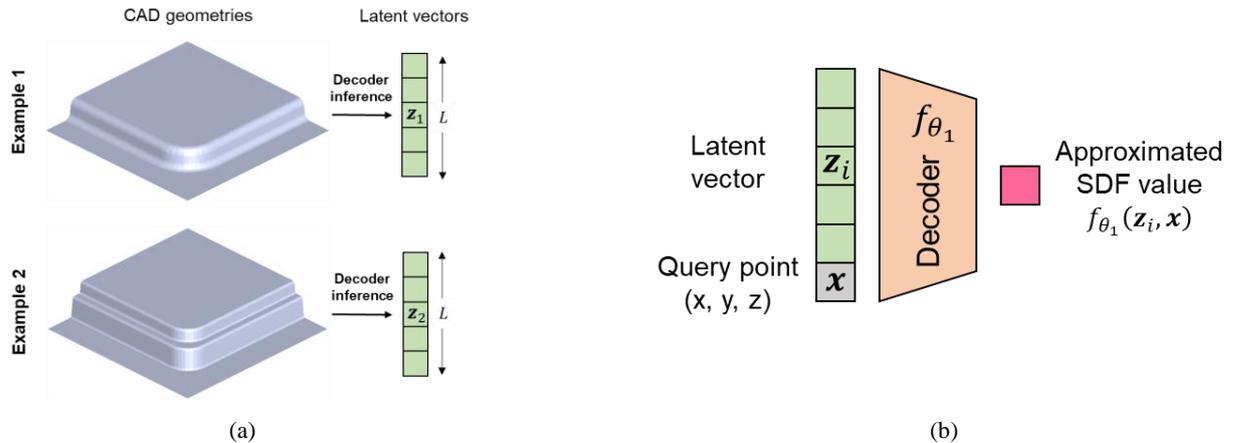

Figure 3 (a) CAD geometries are first converted into low dimensional latent vectors using the decoder inference process. (b) These latent vectors were concatenated with the (x, y, z) coordinates of a 3D query point and the concatenated vector was decoded into an SDF value at the query point by a decoder. The SDF value is conditioned on the latent vector.

## 2.2 Manufacturing performance assessment

The proposed optimisation platform iteratively updates an initial candidate geometry in order to improve its manufacturing performance through a selected stamping process. To perform these updates effectively, a rapid manufacturability assessment must be conducted on a given geometry at each iteration. The use of a surrogate model is proposed here for this purpose. Specifically, a convolutional neural network (CNN) based surrogate model of a hot stamping process recently developed by Attar *et al.* [31] was used. The advantage of adopting this surrogate model here is that it represents an explicit relationship between a candidate geometry and its hot stamping performance. Further, this relationship can be evaluated rapidly, since it does not require the use of FE simulations at each optimisation iteration, and it is differentiable. The differentiability is an important aspect of the optimisation approach adopted here and will be detailed in Section 2.3.

A brief explanation of the manufacturing performance surrogate model is as follows, and further details can be found in the original paper [31]. An overview of the manufacturing performance assessment approach is given in Figure 4. First, a 3D CAD model of a die geometry was projected onto an image which formed its 2D height map, shown in Figure 4(a). This projection was done without information loss since an undercut free geometry is a necessary requirement to avoid collision with stamping tools. Next, the height map was passed through a trained CNN surrogate model of a hot stamping process, $g_{\theta_2}$ with network parameters $\theta_2$. The output of the CNN was the thinning field associated with the input geometry. This thinning field was a projection onto the 2D undeformed blank, shown in Figure 4(b). In this study, the thinning field was used to evaluate the manufacturing performance of the geometry.

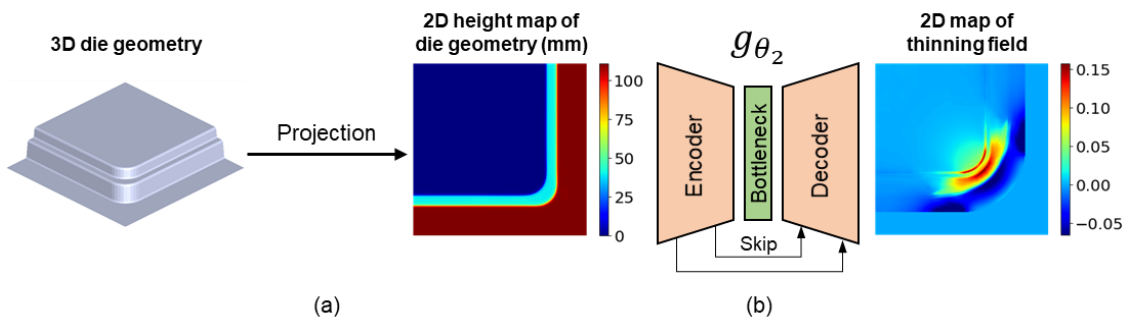

Figure 4 Overview of the manufacturing performance assessment approach. (a) a 3D die geometry was projected onto a 2D plane to form a height map. (b) The height map was passed through the encoder-decoder architecture of a CNN based surrogate model of a hot stamping process, $g_{\theta_2}$, to predict its corresponding thinning field.

## 2.3 Optimisation platform configuration

An overview of the optimisation platform is shown in Figure 5. Fundamentally, the platform consists of three core aspects: 1) an implicit shape representation neural network $f_{\theta_1}$ which maps a latent vector $z$ to a continuous SDF $s$ on a grid of arbitrary resolution, 2) a design evaluator neural network $g_{\theta_2}$ which predicts the manufacturing performance (e.g., thinning or strain fields) of a candidate geometry based on its 2D projected height map, and 3) a differentiable pipeline that bridges the interaction between $f_{\theta_1}$ and $g_{\theta_2}$. By iteratively updating $z$ using a gradient based optimisation technique, a supplied initial geometry is optimised in order to minimise an objective function which is subject to manufacturing constraints. Details about the initialisation required for the platform, as well as forward and backward passes through the platform pipeline are now provided.

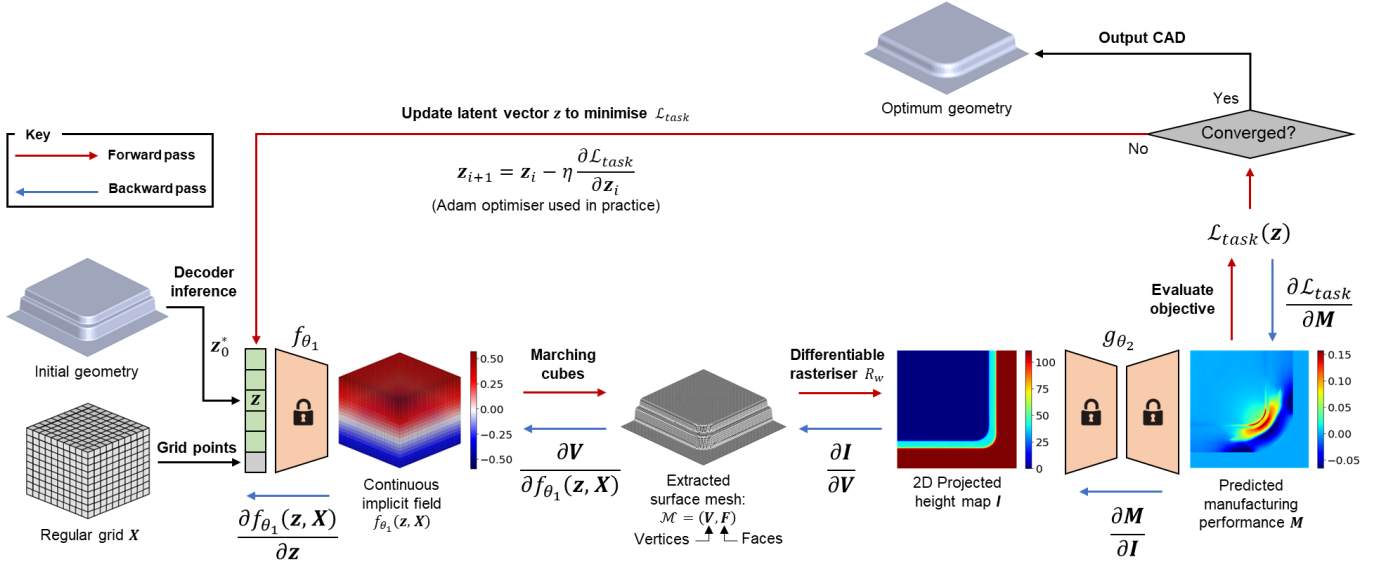

Figure 5 Overview of the optimisation platform pipeline. Black arrows follow operations occurring once for each optimisation process. Red arrows denote forward passes through the optimisation platform which occur once per optimisation iteration. Blue arrows denote backward passes which involve the computation of downstream gradients that are needed for a custom backpropagation and also occur once per optimisation iteration. The custom backpropagation is necessary since it allows the manufacturing performance to drive geometry changes during each optimisation iteration. The padlock symbols on $f_{\theta_1}$ and $g_{\theta_2}$ denote fixed parameters on these neural networks.

### 2.3.1 Pre-optimisation phase

The purpose of the pre-optimisation phase was to perform the initiations that were required for the optimisation platform. The pre-optimisation phase was initiated by supplying an initial CAD geometry which was to be optimised. Decoder inference was conducted on this geometry. This inference process used a trained Auto-Decoder $f_{\theta_1}$ to estimate an optimum latent vector $z_0^*$ which best describes the supplied initial CAD geometry [48]. Decoder inference was in itself an optimisation step and was conducted by backpropagating gradients through the pre-trained network $f_{\theta_1}$. Further details on decoder inference will be given later in Section 4.2. A regular 3D grid $X$ was then assembled and its 3D points $x \in X$ were concatenated with $z_0^*$. This step concluded the pre-optimisation phase. These processes in the pre-optimisation phase occurred once per optimisation case.

### 2.3.2 Forward pass

During each forward pass, a latent vector $z$ and concatenated 3D grid points $x \in X$ were first passed through $f_{\theta_1}$. The result was a 3D SDF $f_{\theta_1}(z, X)$ for the geometry encoded in $z$. The operation performed by $f_{\theta_1}$ can intuitively be thought of as assigning an SDF value to every grid point $x \in X$. The Marching Cubes algorithm, detailed in Appendix A, was then run on the generated SDF and the zero-level-set of the SDF was extracted and converted into an explicit mesh $\mathcal{M} = (V, F)$. The entities $V$ and $F$ define the mesh vertices and faces respectively, and together made up the reconstructed geometry and these were obtained from Marching Cubes. These entities were then passed through a differentiable rasteriser $R_w$ which performed a 3D to 2D orthographic projection of the mesh $\mathcal{M}$, and a 2D projected height map $I$ was obtained. The rasteriser settings $w$ defined camera orientation (e.g., top view) and type of projection (i.e., orthographic or perspective). In practice, the differentiable rasteriser from the Pytorch3D library was used [59].

The manufacturing performance $M$ of the reconstructed geometry was next evaluated. The image based surrogate model $g_{\theta_2}$ described in Section 2.2 was employed for the evaluation. Here, $g_{\theta_2}$ was a CNN based surrogate model of a hot stamping process that was proposed by Attar *et al.* [31]. This surrogate model predicted the thinning field of a candidate geometry based on its height map $I$. These thinning fields were used to define the manufacturing performance $M$. A task-specific objective function $\mathcal{L}_{task}$ was formulated in terms of $M$ and latent vector $z$. The formulation was performed such that minimising $\mathcal{L}_{task}$ with respect to $z$ would improve the manufacturing performance $M$. Further details on this objective function will be given in Section 6.

### 2.3.3 Backward pass

To effectively minimise $\mathcal{L}_{task}$, gradient steps were required at each optimisation iteration since the gradient vector points in the direction of steepest descent. Therefore, iteratively updating $z$ by descending along the $\partial \mathcal{L}_{task}/\partial z$ direction minimised $\mathcal{L}_{task}$ as quickly as possible. Using the chain rule, a custom backward pass was formulated to backpropagate gradients of the objective function $\mathcal{L}_{task}$ to reach $z$, as shown in Equation (5), where · denotes the dot product.

$$\frac{\partial \mathcal{L}_{task}}{\partial z} = \frac{\partial \mathcal{L}_{task}}{\partial M} \cdot \frac{\partial M}{\partial I} \cdot \frac{\partial I}{\partial V} \cdot \frac{\partial V}{\partial f_{\theta_1}(z, X)} \cdot \frac{\partial f_{\theta_1}(z, X)}{\partial z} \tag{5}$$

However, the Marching Cubes algorithm which was used to extract the explicit surface mesh is not differentiable, as demonstrated in recent studies [49,52,60]. This non-differentiability meant that $\partial V/\partial f_{\theta_1}(z, X)$ in Figure 5 was intractable and this created a barrier for computing $\partial \mathcal{L}_{task}/\partial z$ using Equation (5). The non-differentiability of Marching Cubes was recently addressed by Remelli *et al.* [52]. First, Remelli *et al.* queried about how perturbations of the SDF will locally deform the *continuous implicit* surface $S$, which is defined by the zero-level-set of the SDF according to Equation (6).

$$S = \{x \in \mathbb{R}^3 \mid f_{\theta_1}(z, x) = 0\} \tag{6}$$

More formally, the *explicit* mesh vertices $V$ were treated as if they belong to the *implicit* surface $S$ since the surface points are positioned exactly on the implicit surface. This treatment meant that $f_{\theta_1}$ in Equation (5) was evaluated at mesh vertices $V$, rather than at the 3D grid points $X$ when performing the backward pass. Consequently, $f_{\theta_1}(z, X)$ in Equation (5) was replaced by $f_{\theta_1}(z, V)$ and Equation (7) was obtained. If $f_{\theta_1}(z, X)$ were to be used, it would require backpropagating through Marching Cubes, which is non-differentiable, as mentioned above. The final derivative in Equation (5) then became $\partial f_{\theta_1}(z, V)/\partial z$. This derivative can be understood by thinking of the mesh vertices $V$ being a function of $z$ through the implicit surface in Equation (6), rather than through Marching Cubes.

$$\frac{\partial \mathcal{L}_{task}}{\partial z} = \frac{\partial \mathcal{L}_{task}}{\partial M} \cdot \frac{\partial M}{\partial I} \cdot \frac{\partial I}{\partial V} \cdot \frac{\partial V}{\partial f_{\theta_1}(z, V)} \cdot \frac{\partial f_{\theta_1}(z, V)}{\partial z} \tag{7}$$

Using the above treatment, Remelli *et al.* [52] derived an expression for the equivalent to $\partial V/\partial f_{\theta_1}(z, V)$ in their paper. This expression was independent on the method used to extract the *explicit* surface, i.e., Marching Cubes. It is shown by the first equality in Equation (8) and a proof is given in Theorem 2 by Remelli *et al.* [52].

$$\frac{\partial V}{\partial f_{\theta_1}(z, V)} = -\mathcal{N}(V) = -\nabla_V f_{\theta_1}(z, V) \approx -\nabla_V s(V) \tag{8}$$

$\mathcal{N}(V)$ denotes the unit normal vector field at the mesh vertices $V$ and $\nabla_V s(V)$ denotes the spatial gradients of the SDF with respect to $V$. The second equality in Equation (8) comes from Property 2 of SDFs, introduced in Section 2.1.2. The third equality comes from Equation (4), where the neural network $f_{\theta_1}$ is trained to closely approximate the SDF $s$. Here, $\partial V/\partial f_{\theta_1}(z, V)$ can be thought of as the displacement of the implicit surface $S$ due to an infinitesimal perturbation of the implicit field $s$.

Given the theoretical basis described above, Equation (7) was reformulated to facilitate differentiability. First, the gradients of the objective function with respect to mesh vertices $V$ were calculated following Equation (9). In practice, this calculation was done using automatic differentiation in Pytorch by taking the mesh vertices $\boldsymbol{V}$ as leaf nodes in the computational graph [61]. This calculation was possible since the stages upstream of Marching Cubes in Figure 5 (i.e., differentiable rasteriser, $g_{\theta_2}$ and evaluating $\mathcal{L}_{task}$) were all differentiable.

$$\frac{\partial \mathcal{L}_{task}}{\partial \boldsymbol{V}} = \frac{\partial \mathcal{L}_{task}}{\partial \boldsymbol{M}} \cdot \frac{\partial \boldsymbol{M}}{\partial \boldsymbol{I}} \cdot \frac{\partial \boldsymbol{I}}{\partial \boldsymbol{V}} \tag{9}$$

Next, to evaluate Equation (8), the unit normal vector field $\mathcal{N}(\boldsymbol{V})$ was analytically computed by calculating surface gradients $\nabla_{\boldsymbol{V}} f_{\theta_1}(\boldsymbol{z}, \boldsymbol{V})$ at the mesh vertices $\boldsymbol{V}$ via backpropagation through $f_{\theta_1}$. This calculation was again done using automatic differentiation in Pytorch. Finally, substituting Equation (8) and Equation (9) into Equation (7) resulted in Equation (10)

$$\frac{\partial \mathcal{L}_{task}}{\partial \boldsymbol{z}} = \frac{\partial \mathcal{L}_{task}}{\partial f_{\theta_1}(\boldsymbol{z}, \boldsymbol{V})} \cdot \frac{\partial f_{\theta_1}(\boldsymbol{z}, \boldsymbol{V})}{\partial \boldsymbol{z}} \tag{10}$$

where

$$\begin{aligned} \frac{\partial \mathcal{L}_{task}}{\partial f_{\theta_1}(\boldsymbol{z}, \boldsymbol{V})} &= \frac{\partial \mathcal{L}_{task}}{\partial \boldsymbol{V}} \cdot \frac{\partial \boldsymbol{V}}{\partial f_{\theta_1}(\boldsymbol{z}, \boldsymbol{V})} \\ &= -\frac{\partial \mathcal{L}_{task}}{\partial \boldsymbol{V}} \cdot \nabla_{\boldsymbol{V}} f_{\theta_1}(\boldsymbol{z}, \boldsymbol{V}) \end{aligned} \tag{11}$$

To facilitate automatic differentiation with respect to $\boldsymbol{z}$ in practice, Equation (10) was recast as Equation (12). A weighted dot product between $\partial \mathcal{L}_{task}/\partial f_{\theta_1}(\boldsymbol{z}, \boldsymbol{V})$ and $f_{\theta_1}(\boldsymbol{z}, \boldsymbol{V})$ was first taken, and then the differentiation was performed. Here, the gradient $\partial \mathcal{L}_{task}/\partial f_{\theta_1}(\boldsymbol{z}, \boldsymbol{V})$ was pre-computed above and was constant, as labelled in Equation (10). Therefore, performing the differentiation with respect to $\boldsymbol{z}$ in Equation (12) only operated on $f_{\theta_1}(\boldsymbol{z}, \boldsymbol{V})$, and this operation was consistent with Equation (10).

$$\frac{\partial \mathcal{L}_{task}}{\partial \boldsymbol{z}} = \frac{\partial}{\partial \boldsymbol{z}} \left( \frac{1}{|\boldsymbol{V}|} \sum_{v \in \boldsymbol{V}} \alpha_v \frac{\partial \mathcal{L}_{task}}{\partial f_{\theta_1}(\boldsymbol{z}, \boldsymbol{v})} \cdot f_{\theta_1}(\boldsymbol{z}, \boldsymbol{v}) \right) \tag{12}$$

The summation shown in Equation (12) comes from the decomposition of the dot product, as used in the previous equations above. The summation is explicitly shown here to denote the weights $\alpha_v$ which were applied to the gradients of each mesh vertex $v \in \boldsymbol{V}$. These weights were used to ensure that surface points on locally small areas of the geometry (e.g., tight radii) have greater importance. This weighting was required since it is often desirable to optimise these local areas on a stamping geometry over more global features, such as component height [15].

Recall that the geometry was changing after each optimisation iteration. Therefore, the number of mesh vertices was not guaranteed to be consistent between different geometries. For example, a shallow geometry would have less mesh vertices than a deep geometry since it has a smaller surface area. To ensure consistency between geometries at each optimisation iteration, a division by the total number of mesh vertices $|\boldsymbol{V}|$ was introduced in Equation (12).

Equation (12) provided a computationally tractable way to backpropagate the gradient of the objective function $\mathcal{L}_{task}$ all the way to the latent vector $\boldsymbol{z}$. This custom backpropagation ultimately allowed $\partial \mathcal{L}_{task}/\partial \boldsymbol{z}$ to be computed by bypassing the non-differentiable Marching Cubes step, shown in Figure 5. Using $\partial \mathcal{L}_{task}/\partial \boldsymbol{z}$, a gradient step was taken each optimisation iteration $i$, and $\boldsymbol{z}$ was updated to minimise $\mathcal{L}_{task}$. The fundamental gradient step update is illustrated by Equation (13), where $\eta$ is a scalar that controls the step size.

$$\boldsymbol{z}_{i+1} = \boldsymbol{z}_i - \eta \frac{\partial \mathcal{L}_{task}}{\partial \boldsymbol{z}_i} \tag{13}$$

In practice, the Adam optimiser [62] was used to perform this gradient based update, which featured a more complex variant of the fundamental step update shown in Equation (13). This optimiser was selected since it is commonly used in literature for robustly training neural networks by updating their parameters (which have far greater dimensions than *z*) using gradient based techniques.

## 3 Geometry definitions and design of experiments

This section provides the definitions of the geometries which are used in this paper for demonstrating the proposed optimisation platform. Following the definitions, details on design of experiments to generate variants of these geometries are given. These variants were used for training both of the design generator $f_{\theta_1}$ and the manufacturing performance evaluator $g_{\theta_2}$. The remainder of this paper cross-references the geometries presented in this section.

### 3.1 Geometry definitions

Deep drawn box corners were considered as the high level geometry class of interest in this study. These corners are common limiting design features and are found on a range of rectangular or square component designs, e.g., door inners or battery boxes [13,63]. Due to their symmetry, quarter boxes were modelled. Half-lengths of 500 mm and 6° draft angles were used. Blank shapes were defined in accordance with Attar *et al.* [31].

To showcase the non-parametric advantage of the SDF based geometric modelling approach proposed in this study (see Section 2.1), a mix of three geometry subclasses were considered. These subclasses were named standard corners, chamfer corners and stepped sidewalls and could only be described by adopting three different CAD parameterisation schema. These parameterisations are shown in Figure 6(a) and further clarification for how chamfer corners were defined is shown in Figure 6(b). Here, the tool geometries were determined by surface offsets from designed components, and a constant offset of 1.15× blank thickness was used.

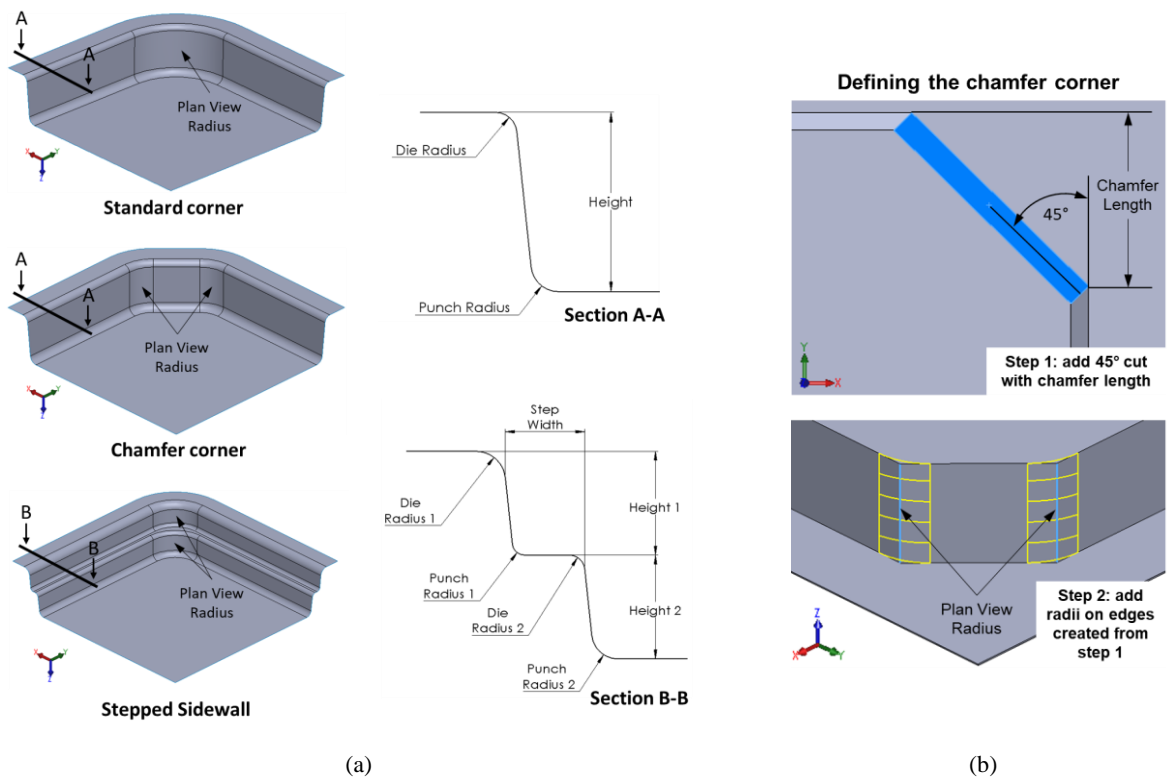

Figure 6 Definition of the three box corner subclasses used in this study: standard corners, chamfer corners and stepped sidewalls. Figure highlights (a) parameterisation of CAD geometries with section views and (b) the two step process used to create chamfer corners.

### 3.2 Design of experiments

To create datasets of geometries, a design of experiments (DoE) was conducted, and variants of the geometries introduced in Section 3.1 were generated. The Latin Hypercube (LHC) DoE technique was used since it is a popular sampling strategy for deterministic computer simulations [64]. Using this technique, an approximately uniform distribution of samples within the design space was obtained while reoccurrences were avoided.

Independent training and testing datasets were created. The training sets were used to train the neural networks $f_{\theta_1}$ and $g_{\theta_2}$. The testing sets were used to test the capability of the networks to make predictions on samples that were not seen during training. Three different LHC runs were conducted for each dataset type, one for each geometry subclass. The training data consisted of 400 samples for each subclass, and the testing data consisted of 100 samples for each subclass. Since three geometry subclasses were considered, the collocated training dataset consisted of 1200 samples and the collocated testing dataset consisted of 300 samples.

The LHC sampling criterion followed the optimisation technique of maximising the minimum distance between points [65]. During this maximisation process, empirically determined linear inequality constraints were imposed. The purpose of these constraints was to preserve geometric integrity during CAD model generation [66,67]. These constraints, along with the considered parameters and ranges for the LHC uniform distributions, are listed in Table 1. The parameters here correspond to those labelled in Figure 6. The CAD model generation was automated due to the relatively large number of samples in the datasets. The automation was achieved using parametric CAD models along with the VBA programming language in SolidWorks.

Table 1 Considered parameters, ranges, and DoE constraints for each parameterisation scheme for the different geometry subclasses

| Parameterisation scheme 1: Standard corners | | | |
|---|---|---|---|
| **Symbol** | **Description** | **Bounds (mm)** | **DoE Constraints** |
| $r_{die}$ | Die radius | 5 – 25 | None |
| $r_{punch}$ | Punch radius | 7.3 – 27.3 | |
| $r_{plan}$ | Plan view radius | 60 – 120 | |
| $H$ | Height | 60 – 120 | |
| **Parameterisation scheme 2: Chamfer corners** | | | |
| **Symbol** | **Description** | **Bounds (mm)** | **DoE Constraints** |
| $r_{die}$ | Die radius | 5 – 25 | $1.7C - R_{plan} + 5.4 > 0$ |
| $r_{punch}$ | Punch radius | 7.3 – 27.3 | |
| $r_{plan}$ | Plan view radius | 60 – 250 | |
| $C$ | Chamfer length | 60 – 140 | |
| $H$ | Height | 60 – 120 | |
| **Parameterisation scheme 3: Stepped sidewalls** | | | |
| **Symbol** | **Description** | **Bounds (mm)** | **DoE Constraints** |
| $r_{die,1}$ | Die radius 1 | 5 – 20 | $SW - r_{punch,1} - r_{die,2} - \epsilon > 0$ |
| $r_{die,2}$ | Die radius 2 | 5 – 20 | $H_1 - r_{die,1} - r_{punch,1} - \epsilon > 0$ |
| $r_{punch,1}$ | Punch radius 1 | 7.3 – 22.3 | $H_2 - r_{die,2} - r_{punch,2} - \epsilon > 0$ |
| $r_{punch,2}$ | Punch radius 2 | 7.3 – 22.3 | $120 - H_1 - H_2 \geq 0$ |
| $r_{plan}$ | Plan view radius | 60 – 120 | $\epsilon = 5$ mm, arbitrary compensation for draft angle |
| $H_1$ | Height 1 | 30 – 100 | |
| $H_2$ | Height 2 | 30 – 100 | |
| $SW$ | Step width | 10 – 50 | |

## 4 Development of implicit neural representations for stamping geometries

This section presents methods that are proposed for learning SDFs of stamping geometries using neural networks. Two approaches for learning these SDFs are introduced. The performances of networks trained using these approaches are evaluated in depth. Design considerations for network training are also presented.

### 4.1 Neural network architecture

Recall from Section 2 that Auto-Decoder neural networks were trained to generate SDFs. The Auto-Decoders used here were based on multi-layer perception architectures. Several architectures were trialled, and these are detailed in Appendix B. The architecture of one of these networks is illustrated in Figure 7. As mentioned previously, the (x, y, z) spatial coordinates of a query point in ambient 3D space was first concatenated with a given latent vector $\boldsymbol{z}$ of length $L$. The concatenated vector formed the input to the network and was of length $L + 3$. This vector was then passed through a series of fully connected layers with activation functions, and each layer outputted an intermediate feature vector with the lengths shown in the figure. A concatenative skip connection was used halfway through the network as this has been reported to improve learning performance [48]. The network output was a scalar SDF value belonging to the input 3D query point. This SDF value was conditioned on the latent vector ***z***.

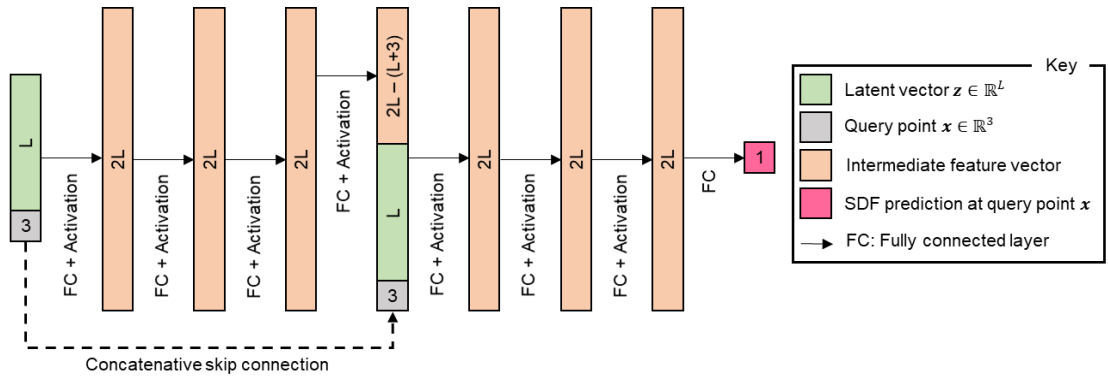

Figure 7 Architecture of an 8-layer Auto-Decoder which was used to generate an SDF value at a 3D query point that was conditioned on a latent vector. The numbers denote vector lengths.

The advantage of adopting an Auto-Decoder is that its formulation is valid for input point clouds of an arbitrary size and distribution. This means that network training and inference can occur with a large number of points densely sampled near to the geometry surface while sparsely sampled far from the surface to allow for better learning performance (see sections below). In addition, a uniform grid can also be accepted as input to produce a continuous volumetric SDF, which is needed during optimisation (see Section 2). Further, this grid can be sampled at arbitrary resolution, which is useful if a particular geometry contains small local features which demand high resolutions. These are major advantages over more traditional Auto-Encoders and variants thereof [68] for the proposed optimisation platform, since their encoders would expect inputs that are similar to the training data.

## 4.2 Approaches for learning SDFs of stamping geometries using neural networks

Inspired by recent computer vision literature on implicit neural representations [48,52,58,60,69], two different approaches for learning SDFs of stamping geometries using neural networks were introduced. So far, all previous work has been focused on learning SDFs for watertight surfaces, for example in [48,55]. However, stamping die geometries are not watertight since they have bounding edges, and this creates a barrier in representing these geometries using SDFs. To overcome this barrier, a new data pre-treatment method is introduced here.

Given a die geometry, which contains geometric features of a component to be stamped, its top surface was first extracted and exported as an STL mesh file, as shown in Figure 8(a). The mesh vertices were scaled such that the die edge was of unit length. The scaled mesh was then subsampled into a point cloud and the subsampling was done using the Trimesh Python library [70]. The configuration of the 3D points for training data depended on the approaches used for learning SDFs, and these approaches are detailed in the following subsubsections.

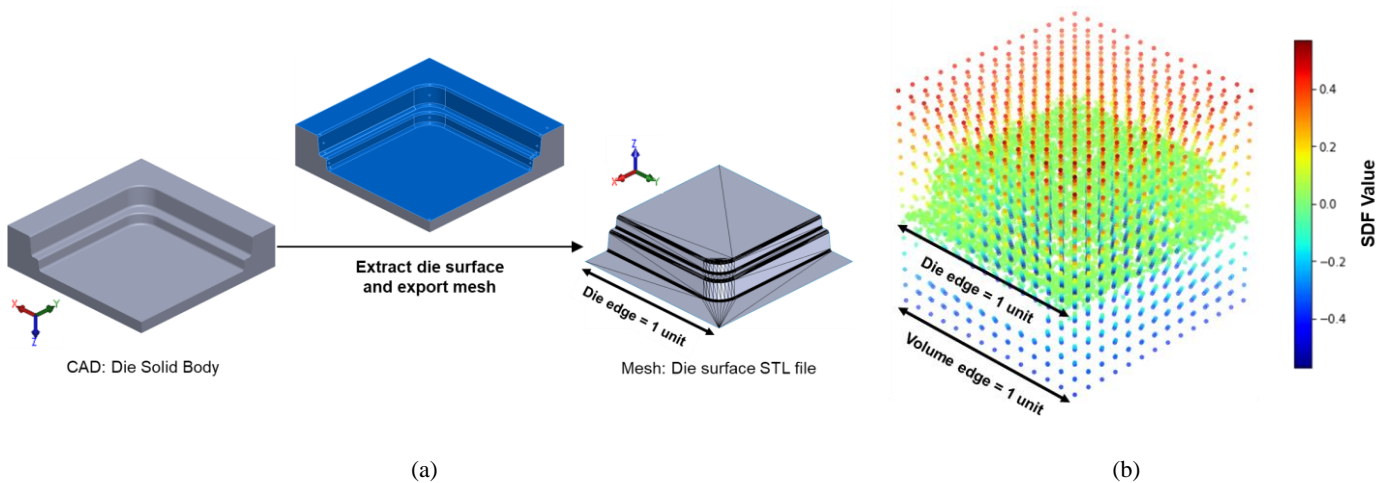

(a)          (b)

Figure 8 Example data-pre-treatment (a) the top surface of a die solid body is extracted and exported as an STL mesh and (b) points are sampled on this surface and within the surrounding cubic volume. Crucially, die edge length equals the cubic volume edge length.

Since SDF values on the surface are expected to be zero, the surrounding volume was also sampled in order to learn a metric SDF. An example sampled point set is shown in Figure 8(b), and each point is given a SDF value for demonstration. In what follows, neural networks were employed to predict the SDF values of these points within the cubic volume of unit edge length for different geometries.

Because the surface had zero thickness, points above the surface had positive values and points below the surface had negative values. To make this formulation possible, the surrounding volume was dimensioned such that its side length was equal to a selected characteristic die edge length, as seen in Figure 8(b). The surface was positioned in the centre of the volume. This positioning allowed approximately equal amounts of positive and negative SDF values to be predicted by the networks which has been reported to support training performance [48].

### 4.2.1 Learning to regress SDF values explicitly (explicit learning approach)

#### a. Approach description

Current work on modelling 3D SDFs using neural networks has mostly focused on learning regressors via explicit supervision of predicted SDF values at training time [48,52,60]. This commonly accepted approach was adopted here to understand to what extent it could be used to model SDFs of geometries for sheet stamping applications. The approach involved a twostep data preparation process; first sampling points and then computing ground truth SDF values for each of these points. This process is now detailed below.

#### b. Data preparation

The first step was to discretise each geometry sample and its surrounding volume into a point set $K$ which contained points in $\mathbb{R}^3$. A breakdown of the adopted point sampling strategy is demonstrated in Figure 9 for one geometry. Points were sampled more aggressively in the space near the surface to capture a more detailed SDF around the surface. To do this effectively, 3000 points were first sampled exactly on the geometry surface, where an SDF value of zero is expected. Then, two noisy point sets were further generated, one with low noise and the other with high noise. These sets were generated by perturbing the positions of the surface points by zero mean Gaussian distributions with low and high standard deviations respectively, as shown in Figure 9. To cover the remainder of the domain, a uniform unit cube with 3000 points was generated. These point sets were collated which resulted in $|K| = 12000$ total points for one geometry sample. This point sampling process was repeated for all geometries in the training and testing datasets.

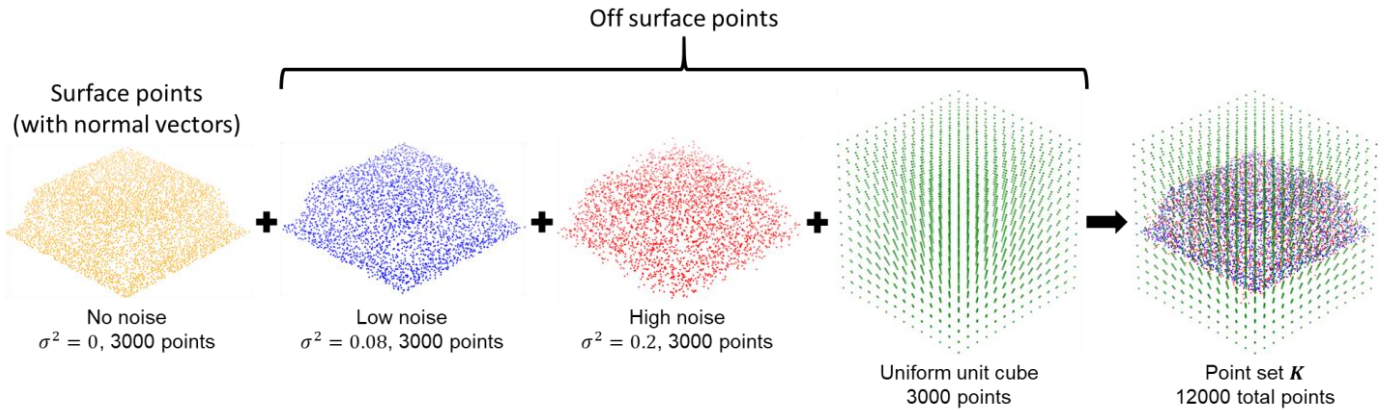

Figure 9 Breakdown of point subsampling strategy used for explicit learning of SDFs. A total point set is collected, which consists of both points on and off the surface. Here, $\sigma$ denotes the standard deviation of a zero mean Gaussian distribution added as noise.

The second step was to analytically compute the SDF values at each of the subsampled points for all geometry samples. For each geometry, the surface points were assigned an SDF value of zero, and their position vectors were stored in a KD-tree. For each off surface point $x_o$, its nearest surface point $x_s$ was identified by querying the KD-tree, and the distance between them was computed by taking the Euclidian norm of their difference. The sign was determined by the sign of the dot product between the normal vector of the surface point $n_s$ and their difference, i.e., $\text{sign}(n_s \cdot (x_o - x_s))$.

The positions of the 3D points were taken as network inputs and their analytical SDF values were taken as ground truth targets for this learning approach. The dataset $D^i$ for the $i$th geometry is summarised by the set shown in Equation (14).

$$D^i = \{(x_j, s_j) : \ s_j = \text{SDF}^i(x_j)\} \tag{14}$$

where for each point $x_j$ there was a corresponding $s_j$. The complete dataset $D^i{}_{i=1}^N$ consisted of $N$ geometry samples represented with analytical signed distance functions $\text{SDF}^i{}_{i=1}^N$ evaluated at each subsampled point $x_j$.

### c. Training and inference

To supervise the SDF predictions explicitly, the clamped $L_1$ loss function was used as suggested by Park *et al.* [48] and is shown in Equation (15).

$$\mathcal{L}(f_{\theta_1}(z,x),s) = |\text{clamp}(f_{\theta_1}(z,x),\delta) - \text{clamp}(s,\delta)| \tag{15}$$

In this equation, $s$ and $f_{\theta_1}(z,x)$ are the ground truth and predicted SDF values at point $x$ respectively. The function $\text{clamp}(\varepsilon,\delta) := \min(\delta, \max(-\delta, \varepsilon))$ introduces the parameter $\delta > 0$, known as the truncation distance, to control the distance from the surface over which the network is expected to learn the SDF. In this study, $\delta = 0.05$ was used, and this value was inspired from ablation studies performed by Park *et al.* [48]. This loss function concentrated the networks learning capability on details close to the zero-level set of the SDF.

Using this loss function, an objective function to be minimised at training time was formulated in terms of all geometry samples in a training batch with batch size $B$, show in Equation (16).

$$\arg\min_{\theta_1,\{z_i\}_{i=1}^B} \frac{1}{B} \sum_{i=1}^B \mathcal{L}_{training}^E(\theta_1, z_i) = \arg\min_{\theta_1,\{z_i\}_{i=1}^B} \frac{1}{B} \sum_{i=1}^B \left( \frac{1}{|K|} \sum_{x_j \in K} \mathcal{L}(f_{\theta_1}(z_i, x_j), s_j) + \lambda \|z_i\|_2 \right) \tag{16}$$

where the superscript $E$ denotes this explicit learning approach described. The minimisation was performed with respect to both the network parameters $\theta_1$ and the latent vectors for geometries in the training batch $\{z_i\}_{i=1}^B$. The addition of the regularisation term weighted by $\lambda$ ensured that the latent vector magnitudes were concentred near the origin. This concentration, which has been reported to help training and inference convergence [48], enabled similar shapes to have similar latent vectors. As recommended by Park *et al.* [48], all latent vectors were initialised from a zero mean Gaussian distribution with low standard deviation $N(0, 0.01^2)$ to further ensure that similar shapes were represented by similar latent vectors.

Equation (16) was used to train $f_{\theta_1}$ to determine the network parameters $\theta_1$ using the explicit learning approach. Minimising Equation (16) with respect to both $\theta_1$ and $\{z_i\}_{i=1}^B$ at training time enabled the latent vector of an unseen geometry to be inferred using the trained decoder $f_{\theta_1}$. At inference time, the network parameters $\theta_1$ were fixed and an optimum latent vector for an unseen geometry was inferred through an iterative optimisation process. This decoder inference process involved minimising the expression shown in Equation (17) which was conducted iteratively by backpropagating through the trained decoder at each iteration.

$$\arg\min_z \mathcal{L}_{inference}^E(z) = \arg\min_z \frac{1}{|K|} \sum_{x_j \in K} \mathcal{L}(f_{\theta_1}(z, x_j), s_j) + \lambda \|z\|_2 \tag{17}$$

A graphical representation for the iterative decoder inference process using this explicit learning approach is shown in Figure 10. Given an unseen geometry, a point set $K$ which contained points in $\mathbb{R}^3$ was sampled and ground truth SDF values at these points were pre-computed in the same way as was done for the training data. Next, a random vector $z_0 \sim N(0, 0.01^2)$ of length $L$ was sampled from a zero mean Gaussian distribution and this was the initial latent vector to be optimised. For each iteration, the latent vector $z$ was repeated for $|K|$ instances, and the coordinates of each 3D point were concatenated to each instance to form a tensor of size $(L + 3) \times |K|$. This tensor was passed through the trained decoder $f_{\theta_1}$ to predict SDF values at each of the points. A comparison with ground truth SDF values was performed and the loss $\mathcal{L}_{inference}^E(z)$ was computed following Equation (17). Using the gradient of this loss, the latent vector $z$ was iteratively updated by the Adam optimiser to minimise the loss until convergence. At convergence, an optimum latent vector $z^*$ was obtained, which best described the unseen geometry using this approach.

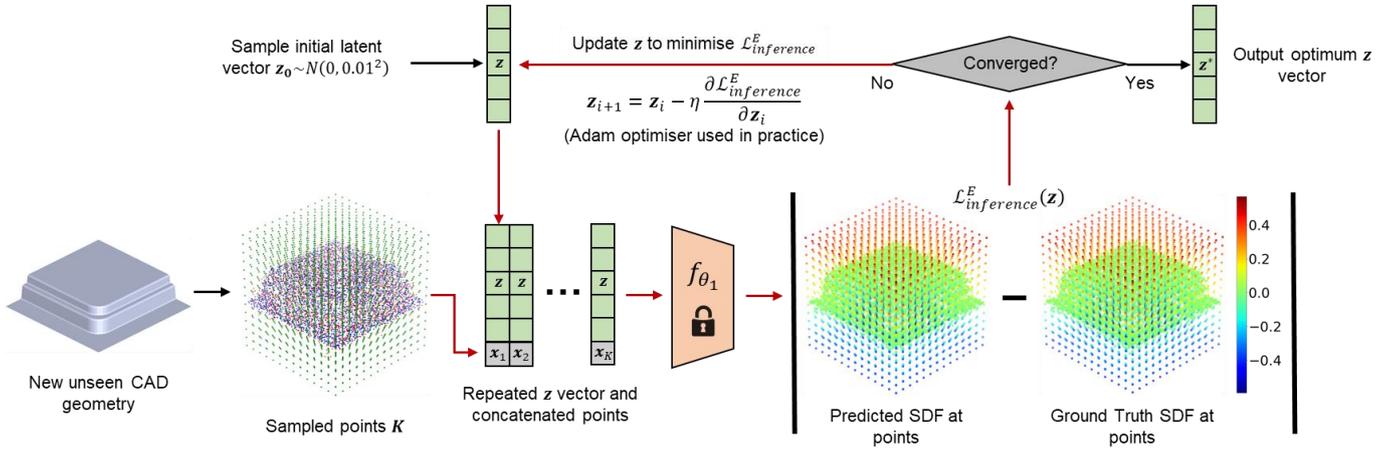

Figure 10 Decoder inference process to obtain a latent vector representation of an unseen geometry using a trained network $f_{\theta_1}$ which was trained using the explicit learning approach. Red arrows occur once per inference iteration.

### 4.2.2 Learning geometric properties of SDFs (implicit learning approach)

*a. Approach description*

The implicit learning approach did not require explicit supervision of SDF values at training time. Instead, the key idea was to learn SDFs implicitly by exploiting and supervising their geometric properties. By learning the intrinsic relationship between the coordinates of points in $\mathbb{R}^3$ and the geometric properties that all ground truth SDFs obey by definition, this approach was expected to provide superior performance over learning a naive regressor (i.e., explicit approach). This approach was inspired by recent work in learning high quality scene and 3D objects reconstruction, such as rooms, tables and chairs with thin features, for computer vision applications [55,58,69].

Recall the definition and geometric properties of SDFs from Section 2.1. By definition, the SDF values of points that lie exactly on the geometry surface are zero. The magnitude of spatial gradient vectors of SDFs are equal to one and these vectors at the geometry surface align with the surface normals. Together, the definition and properties describe attributes of SDFs that can be classified into two categories: off the geometry surface and on the surface. The dataset and loss function for this approach were prepared by considering these two attribute categories.

*b. Data preparation*

To prepare the dataset for this approach, points in $\mathbb{R}^3$ were subsampled on and around all geometry samples and this step was similar to the one used for the explicit learning approach. However, here the off surface and on surface points were kept separate and stored as two point sets to facilitate learning SDF attributes for these two categories. A breakdown of this point subsampling strategy is shown in Figure 11 for one geometry sample. 9000 total off surface points were subsampled according to the figure, and an additional 9000 surface points were subsampled and stored separately. The surface points also came with normal vectors as part of the subsampling process. This point subsampling process was repeated for all geometries in the training and testing datasets. The deviation from the attributes of SDFs was penalised solely based on the predicted SDF values. Therefore, this approach did not need ground truth SDF samples and can be considered as self-supervised. Details on how this self-supervised approach was employed are provided in the following.

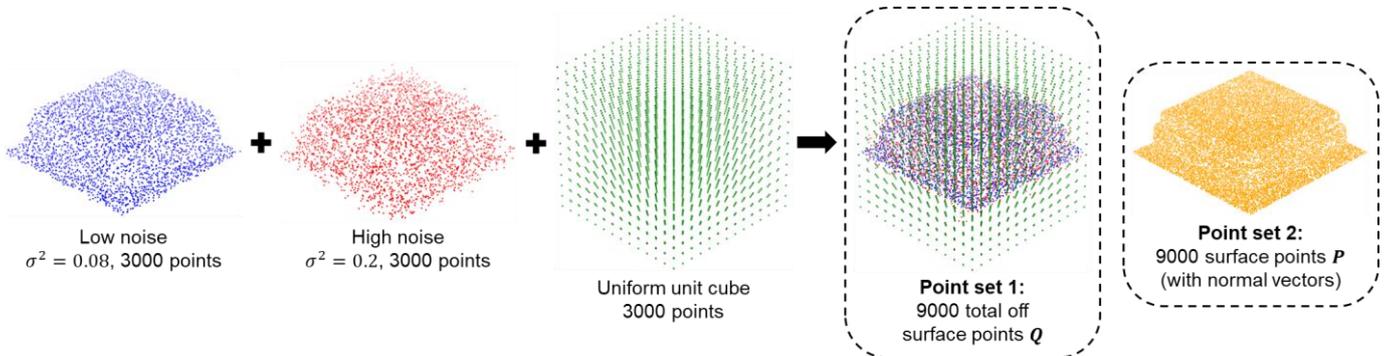

Figure 11 Breakdown of point subsampling strategy used for implicit learning of SDFs. Two point sets were sampled: off surface points, and surface points with surface normal vectors. Here, $\sigma$ denotes the standard deviation of a zero mean Gaussian distribution added as noise for off surface points close to the surface.

*c. Training and inference*

To effectively learn the attributes of SDFs, a new loss function to be minimised was cast in terms of the aforementioned point categories, surface, and off surface points, as shown in Equation (18). As before, the loss was formulated in terms of all geometry samples in a training set batch with batch size $B$. The minimisation was performed with respect to both the network parameters $\theta_1$ and the latent vectors for geometries in the training batch $\{z_i\}_{i=1}^{B}$. The three terms in Equation (18) are separately explained in detail in the following paragraphs.

$$\arg\min_{\theta_1,\{z_i\}_{i=1}^{B}} \frac{1}{B} \sum_{i=1}^{B} \mathcal{L}_{training}^{I}(\theta_1, z_i) = \arg\min_{\theta_1,\{z_i\}_{i=1}^{B}} \frac{1}{B} \sum_{i=1}^{B} \left( \mathcal{L}_{surface}(\theta_1, z_i) + \mathcal{L}_{off\ surface}(\theta_1, z_i) + \mathcal{L}_{reg}(z_i) \right) \quad (18)$$

The first term in Equation (18) contained a loss concerned with the behaviour of the predicted SDF at the geometry surface $\mathcal{L}_{surface}(\theta_1, z_i)$, and this term is shown in Equation (19).

$$\mathcal{L}_{surface}(\theta_1, z_i) = \frac{1}{|P|} \sum_{x_j \in P} \left( \lambda_1 |f_{\theta_1}(z_i, x_j)| + \lambda_2 \alpha_j \left\| \nabla_{x_j} f_{\theta_1}(z_i, x_j) - n_j \right\|_2 \right) \quad (19)$$

This surface loss ensured supervision of surface attributes by penalising deviations from them. This loss was formulated on the surface point set $P$ and contained two terms, weighted by the scalars $\lambda_1$ and $\lambda_2$ respectively. Minimising the first term ensured that the magnitude of an SDF prediction $f_{\theta_1}(z_i, x_j)$ for point $x_j \in P$ on the surface of the geometry encoded in $z_i$ tended toward zero, which satisfies the definition of SDFs. The second term ensured that the spatial gradient vector of the predicted SDF at point $x_j \in P$ on the surface $\nabla_{x_j} f_{\theta_1}(z_i, x_j)$ aligned with the unit normal vector at that point $n_j$. The $L_2$ norm $\|\cdot\|_2$ was taken between the two to allow alignment in both vector direction and magnitude.

So far, the importance of certain geometric features on manufacturing performance has not been considered during network training. For stamping applications, fillet radii are geometric features which are highly influential in determining manufacturing performance, as explained in the introduction. These features are small in scale when considered in the context of the global component geometry. This means that a surface subsampled with an oriented point cloud as shown in Point set 2 in Figure 11 would have relatively few points on these small radii regions.

To tackle this imbalance and reflect the importance of small local geometric features in Equation (19), an additional scaling factor $\alpha_j$ was introduced which operated in a point-wise manner. The purpose of this scaling factor was to give more weighting to small scale regions on the geometry surface where accurate reconstructions are crucial, for example fillet radii as described above. The value of $\alpha_j$ was set equal to a scalar $\beta > 1$ if point $x_j$ met the conditions set in Equation (20)

$$\alpha_j = \begin{cases} \beta, & \langle n_j \cdot n_{ref} \rangle < \tau \\ 1, & \langle n_j \cdot n_{ref} \rangle \geq \tau \end{cases} \quad (20)$$

where $n_{ref}$ is a unit vector pointing in a reference direction and $n_j$ is the unit normal vector of point $x_j$. The $\langle \cdot \rangle$ notation denotes the cosine similarity and $\tau \in [0,1]$ is the cosine similarity allowance. These conditions were designed to automatically identify regions of interest which contained points to be given a higher weighting. In this study, $n_{ref}$ was set equal to the basis vector pointing in the z direction $n_z = [0,0,1]$, $\tau = 1$ and $\beta = 3$. These settings meant that the second term in Equation (19) was scaled by a factor of 3 if point $x_j \in P$ did not lie on flat horizontal surfaces. Figure 12 graphically illustrates regions which contain points influenced by this scaling. In the figure, the cones show the surface normal vector at each subsampled point on the geometry surface. The red zones illustrate identified fillet and sidewall regions to be given a scaling in Equation (19) and the green zones illustrate unscaled horizontal surfaces. This scaling gave more importance to aligning normals that lie on locally curved areas such as radii, as well as the sidewalls.

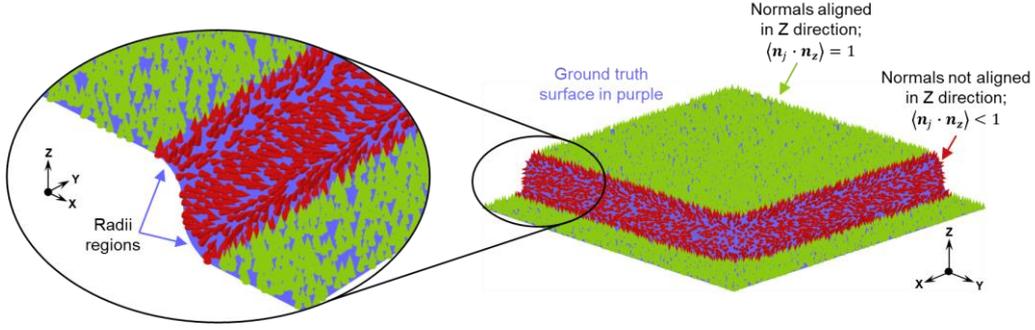

Figure 12 Cone plots showing surface normal vectors pointing away from the surface. Cones are stemming from every surface point subsample. Normals on flat surfaces align with the $z$ direction and are shown in green. Normals on radii and sidewall regions shown in red.

The second term in Equation (18) contained a loss concerned with the behaviour of the predicted SDF away from the geometry surface $\mathcal{L}_{off\ surface}(\theta_1, \mathbf{z}_i)$, shown in Equation (21).

$$\mathcal{L}_{off\ surface}(\theta_1, \mathbf{z}_i) = \frac{\lambda_3}{|\mathbf{Q}|} \sum_{\mathbf{x}_k \in \mathbf{Q}} \left( \left\| \nabla_{\mathbf{x}_k} f_{\theta_1}(\mathbf{z}_i, \mathbf{x}_k) \right\|_2 - 1 \right)^2 \qquad (21)$$

Minimising this term ensured that the spatial gradient vector of the predicted SDF at point $\mathbf{x}_k \in \mathbf{Q}$ away from the surface $\nabla_{\mathbf{x}_k} f_{\theta_1}(\mathbf{z}_i, \mathbf{x}_k)$ had a unit magnitude, which satisfied the solution to the Eikonal equation (see Equation (2)). The magnitude of the spatial gradient vector was computed by taking the $L_2$ norm $\|\cdot\|_2$.

The final term in Equation (18) was a regularisation term that was added for the same reasons mentioned in Section 4.2.1, shown in Equation (22). The $\lambda$ terms in the presented equations denoted scalar weights that were manually tuned to balance the magnitudes of each of the terms that make up Equation (18).

$$\mathcal{L}_{reg}(\mathbf{z}_i) = \lambda_4 \|\mathbf{z}_i\|_2 \qquad (22)$$

Equation (18) was used to train $f_{\theta_1}$ to determine the network parameters $\theta_1$ using the implicit learning approach. At inference time, similar to the explicit learning approach, the network parameters $\theta_1$ were fixed and an optimum latent vector for an unseen geometry was inferred through an iterative optimisation process. This decoder inference process involved minimising the expression shown in Equation (23) which was conducted iteratively by backpropagating through the trained decoder at each iteration.

$$\arg\min_{\mathbf{z}} \mathcal{L}^I_{inference}(\mathbf{z}) = \arg\min_{\mathbf{z}} \frac{1}{|\mathbf{P}|} \sum_{\mathbf{x}_j \in \mathbf{P}} \left( \lambda_1 |f_{\theta_1}(\mathbf{z}, \mathbf{x}_j)| + \lambda_2 \alpha_j \left\| \nabla_{\mathbf{x}_j} f_{\theta_1}(\mathbf{z}, \mathbf{x}_j) - \mathbf{n}_j \right\|_2 \right) \\ + \frac{\lambda_3}{|\mathbf{Q}|} \sum_{\mathbf{x}_k \in \mathbf{Q}} \left( \left\| \nabla_{\mathbf{x}_k} f_{\theta_1}(\mathbf{z}, \mathbf{x}_k) \right\|_2 - 1 \right)^2 + \lambda_4 \|\mathbf{z}\|_2 \qquad (23)$$

A graphical representation for the iterative decoder inference process using this implicit learning approach is shown in Figure 13. An unseen geometry was first sampled into two point sets as explained above for the training data. Similar to the explicit learning approach, a random vector $\mathbf{z}_0 \sim N(0, 0.01^2)$ of length $L$ was sampled from a zero mean Gaussian distribution and this was the initial latent vector to be optimised. For each iteration, and for each of the two point sets, the latent vector $\mathbf{z}$ was repeated for as many instances as there were points in each point set. The coordinates of each 3D point were concatenated to each instance to form tensors of size $(L + 3) \times |\mathbf{P}|$ and $(L + 3) \times |\mathbf{Q}|$ for the surface and off surface point sets respectively. These tensors were passed through the trained decoder $f_{\theta_1}$ and SDF values were predicted at each point in the two point sets. The spatial gradients of these predictions were then taken as necessary and the loss function $\mathcal{L}^I_{inference}(\mathbf{z})$ in Equation (23) was computed. Using the gradient of this loss, the latent vector $\mathbf{z}$ was iteratively updated by the Adam optimiser to minimise the loss until convergence. At convergence, an optimum latent vector $\mathbf{z}^*$ was obtained, which best described the unseen geometry using this approach.

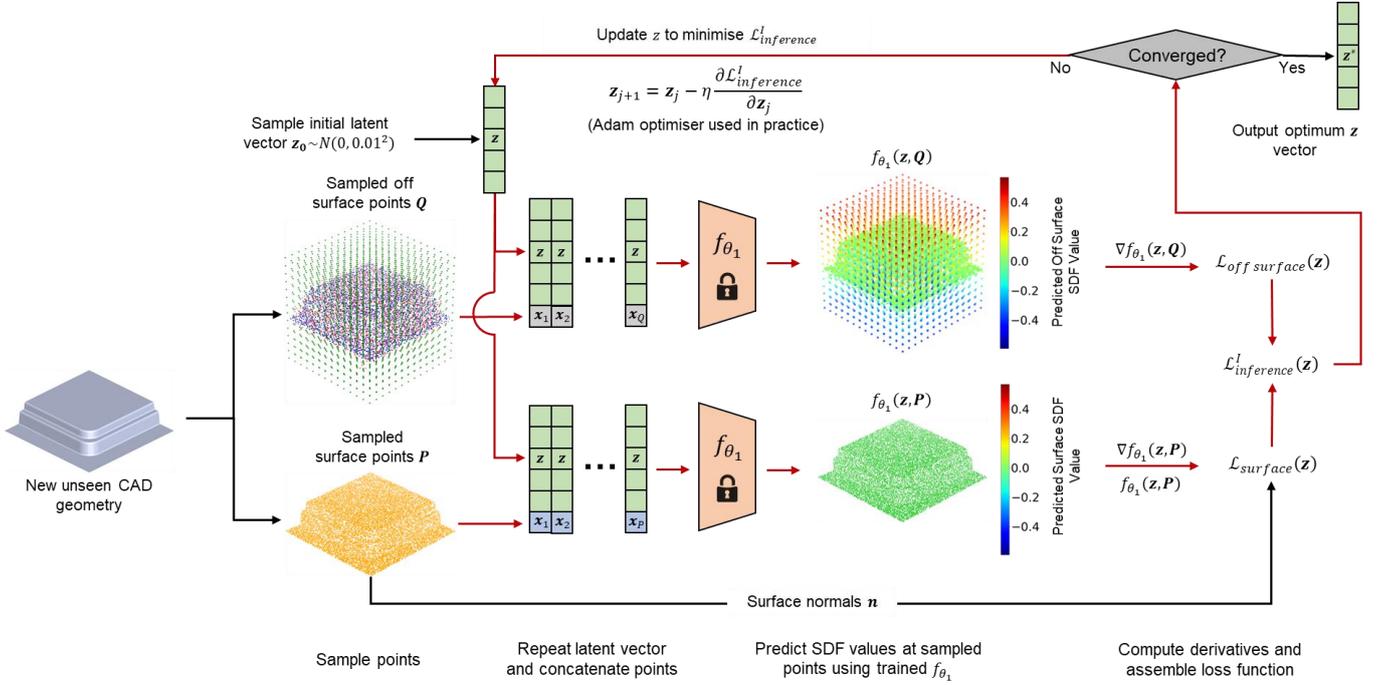

Figure 13 Decoder inference process to obtain a latent vector representation of an unseen geometry using a network $f_{\theta_1}$ using the implicit learning approach. Red arrows occur once per inference iteration.

## 4.3 Design considerations for training and training history

### 4.3.1 Training settings

The Auto-Decoder $f_{\theta_1}$ introduced in Section 4.1 was trained using both training approaches introduced in Section 4.2. For the explicit learning approach, the ReLU activation function was used as recommended by Park *et al.* [48]. For the implicit learning approach, the SoftPlus activation function was used since it is more differentiable than the commonly used ReLU function. This enhanced differentiability was conjectured to be useful for the implicit learning approach since computing the loss function first involved computing gradients of the predicted SDFs. These gradients were computed via backpropagation through $f_{\theta_1}$ using automatic differentiation [61].

For both approaches, the network parameters $\theta_1$ and latent vectors $\{z_i\}_{i=1}^{B}$ were iteratively updated during training and these updates occurred after each training batch. During this iterative process, the optimiser sought to find the combinations of $\theta_1$ and $\{z_i\}_{i=1}^{B}$ that best minimised the loss function for each training approach. In this way, the networks were able to learn compact functions that could predict SDFs of families of box corner geometries. The networks were trained in the PyTorch framework using the commonly recommended Adam optimiser [62] with default beta parameters of $\beta_1 = 0.9$ and $\beta_2 = 0.999$. An initial learning rate of $1 \times 10^{-4}$ was used and this was halved every 500 epochs up to a minimum of $5 \times 10^{-6}$. These learning rates were the same for both $\theta_1$ and $\{z_i\}_{i=1}^{B}$ updates. All parameters in the loss functions for both approaches were determined empirically and are summarised in Table 2.

Table 2 Empirically determined loss function parameters for explicit and implicit training approaches

| Training approach | Parameter | Symbol | Value |
|---|---|---|---|
| Explicit | SDF truncation distance | $\delta$ | 0.05 |
| | Latent vector regularisation loss term weight | $\lambda$ | $1 \times 10^{-4}$ |
| Implicit | Surface SDF loss term weight | $\lambda_1$ | 3 |
| | Normals loss term weight | $\lambda_2$ | 0.7 |
| | Eikonal loss term weight | $\lambda_3$ | 0.9 |
| | Latent vector regularisation loss term weight | $\lambda_4$ | $1 \times 10^{-4}$ |
| | Point weight for normals loss term | $\beta$ | 3 |
| | Cosine similarity allowance | $\tau$ | 1 |

### 4.3.2 Dataset ordering

Gradient updates by the Adam optimiser at training time occurred after each training batch, and each training batch was presented to the networks sequentially during training. Since geometries from different subclasses were considered (see Section 3.1), careful attention was paid to ordering the data samples in the training dataset. The training data was ordered such that each batch contained an equal number of geometry samples from each geometry subclass. This was achieved by periodically arranging samples from each geometry subclass in the dataset, as shown in Figure 14. Further, since there were 3 subclasses in total, the batch size which chosen to be a multiple of 3. Here, a batch size of $B = 21$ geometries was selected. This selection ensured gradient updates contained information on 7 random (i.e., random because of Latin Hypercube sampling) samples from each of the 3 subclasses. This batch size allowed learning SDFs from all subclasses uniformly while being small enough to efficiently fit in GPU memory.

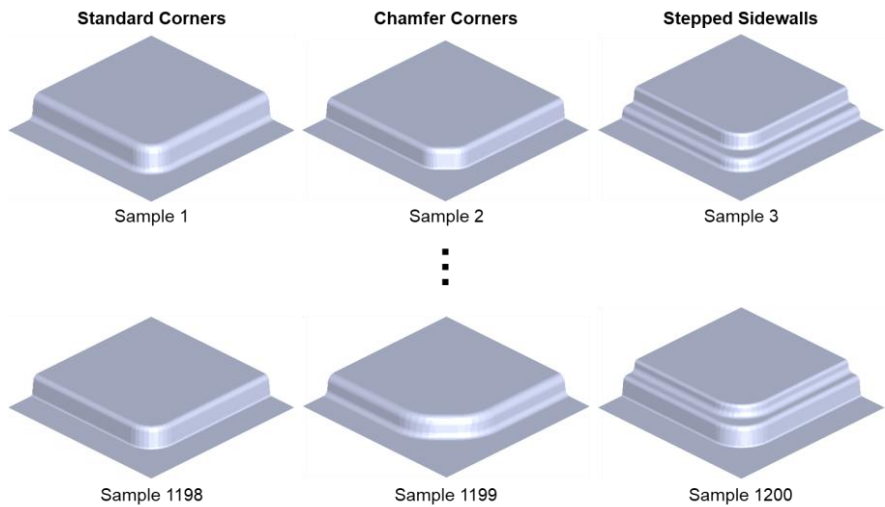

Figure 14 Ordering of geometries in the training dataset; random geometries from all 3 geometry subclasses were arranged periodically. Note that CAD geometries are shown here for clarity whereas the training dataset contained point clouds sampled from these geometries.

### 4.3.3 Training history

Figure 15 shows the training loss histories for both the explicit and implicit learning training approaches. The curve in Figure 15(a) belongs to Explicit Net 1 (given in Appendix A.2) and is representative of all networks trained using the explicit learning approach. The steady declines in the loss values provide evidence of good training stability for both approaches. Training for both approaches was left to run for 24 hours and was run on an NVIDIA P100 GPU in the Google Colab environment. In this timeframe, the explicit learning approach completed 8000 epochs (i.e., 8000 complete iterations of the training dataset), while the implicit approach completed 3800 epochs. The implicit approach completed less epochs in the same 24 hour training timeframe because spatial gradients of the SDF predictions had to be calculated before assembling the loss function at each batch iteration, as explained earlier. However, convergence was seen to occur at approximately 1500 epochs for the implicit approach, while the explicit approach took approximately 7000 epochs. A further study on the effect of the number of subsampling points and number of geometries in the training dataset on the SDF generation accuracy and training time was outside the scope of this work. Nevertheless, it is expected that significant training speeds ups could be achieved by using fewer subsampling points and geometries but may compromise performance accuracy.

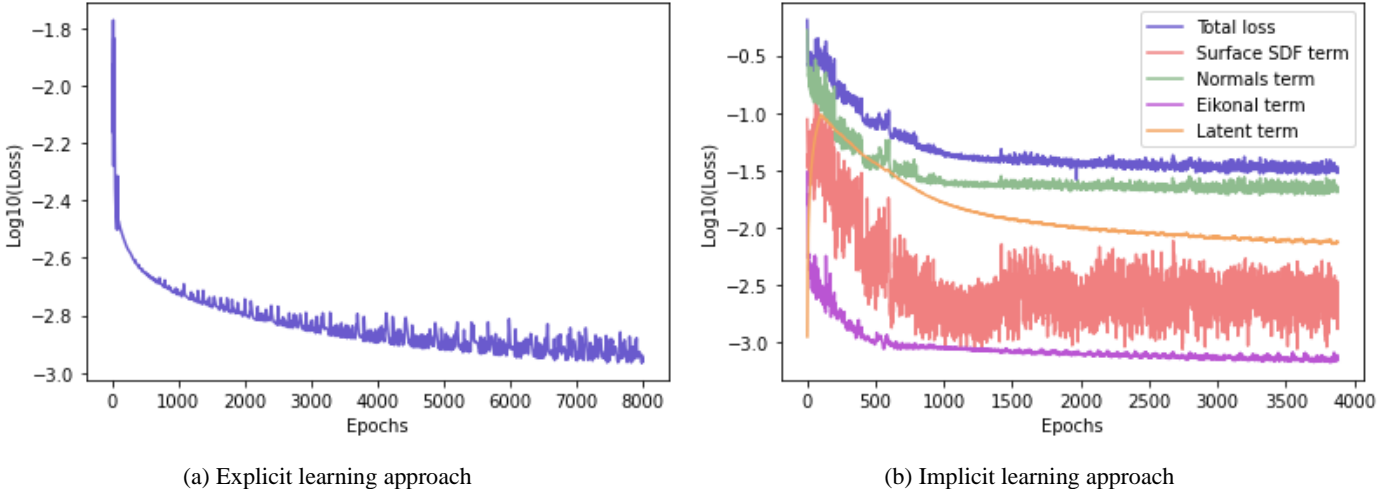

(a) Explicit learning approach    (b) Implicit learning approach

Figure 15 Loss histories for the proposed neural networks trained using the two learning approaches. The plots show (a) the history of Equation (16) and (b) the history of Equation (18). Since the loss of the implicit learning approach is more complex, histories of each loss term that made up the total loss are also presented.

## 4.4 Quantitative evaluation of shape representation performance

After successfully training the Auto-Decoder networks $f_{\theta_1}$ using both of the introduced training approaches, their performances were first evaluated quantitatively. For this evaluation, similarity metrics were used to measure the difference between geometries reconstructed from $f_{\theta_1}$ and the ground truth geometries exported from CAD software. These metrics were inspired by recent literature on implicit neural representations [47,48,58] and were Chamfer Distance, Hausdorff Distance and Mesh Similarity and are defined below.

### 4.4.1 Performance metrics definition

The distance metrics, Chamfer and Hausdorff, required the continuous geometry surfaces to be first discretised into point sets $P_{GT}$ and $P_R$ for ground truth and reconstructed geometries respectively. The Mesh Similarity metric required $P_{GT}$ and the reconstructed surface mesh $\mathcal{M}$. To generate $P_{GT}$, points were sampled on the surface of the STL mesh file that was exported from CAD software. To generate $P_R$, a forward pass was first required through $f_{\theta_1}$ to generate the continuous SDF. Then, the marching cubes algorithm was used to extract the reconstructed surface mesh $\mathcal{M}$. Finally, points were sampled on $\mathcal{M}$. A generous 30,000 points were sampled for both $|P_{GT}|$ and $|P_R|$ to effectively approximate the continuous surfaces. The point set generating process is depicted in the black rectangles in Figure 16, and what's needed for computing each similarity metric is shown in the green rectangle.

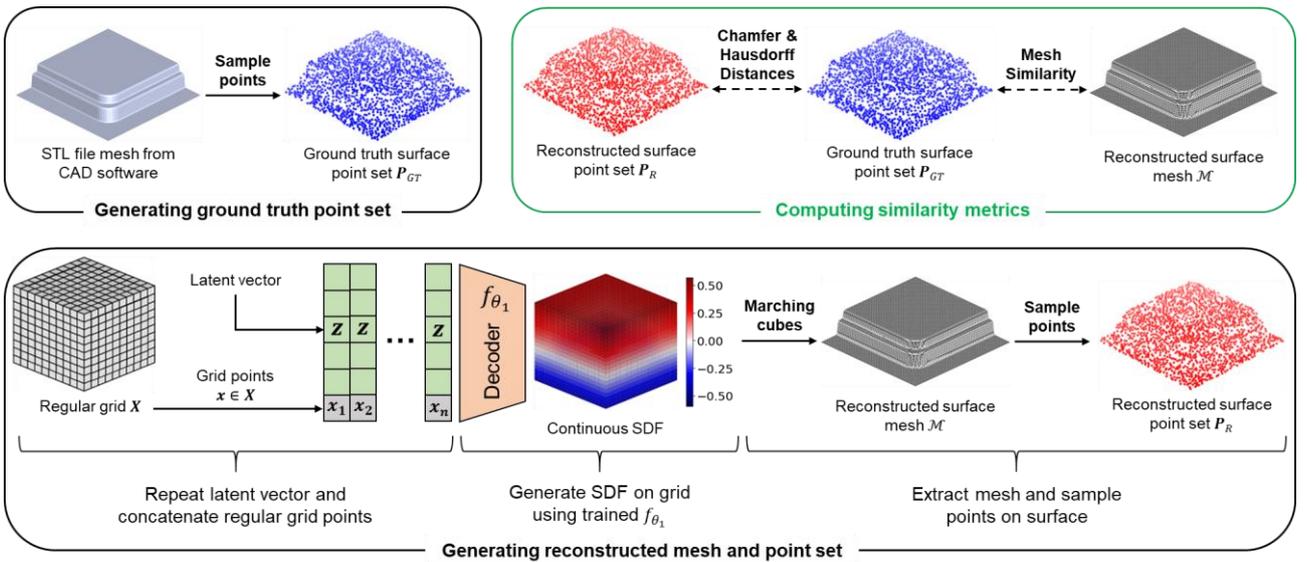

Figure 16 Generating data (black rectangles) to compute similarity metrics (green rectangle). This process was repeated for all geometries in the training and testing datasets.

*a. Chamfer Distance*

The Chamfer Distance measured the difference between reconstruction and ground truth surfaces by considering the distances of all sampled points. Given the two pre-computed point sets $\boldsymbol{P}_{GT}$ and $\boldsymbol{P}_R$, this metric was computed by taking the sum of the squared distances between each point in one set and its nearest in the other set using Equation (24).

$$d_C(\boldsymbol{P}_{GT}, \boldsymbol{P}_R) = \sum_{x_i \in P_{GT}} \min_{x_j \in P_R} \|x_i - x_j\|_2^2 + \sum_{x_i \in P_{GT}} \min_{x_j \in P_R} \|x_i - x_j\|_2^2 \qquad (24)$$

The two terms being summed are each the one sided chamfer distances and summing ensures the distance is symmetric $d_C(\boldsymbol{P}_{GT}, \boldsymbol{P}_R) = d_C(\boldsymbol{P}_R, \boldsymbol{P}_{GT})$. The nearest distances were computed efficiently by the use of a KD-tree. The lower the Chamfer Distance, the better the reconstruction performance.

*b. Hausdorff Distance*

The Hausdorff Distance measured the maximum distance between reconstructions and ground truth surfaces. Given the two pre-computed point sets $\boldsymbol{P}_{GT}$ and $\boldsymbol{P}_R$, this metric was computed by taking the maximum distance between any point in one set and its nearest in the other set using Equation (25).

$$d_H(\boldsymbol{P}_{GT}, \boldsymbol{P}_R) = \max\left(\max_{x_i \in P_{GT}} \min_{x_j \in P_R} \|x_i - x_j\|_2, \max_{x_j \in P_R} \min_{x_i \in P_{GT}} \|x_i - x_j\|_2\right) \qquad (25)$$

The two terms in the bracket are each the one sided Hausdorff distances and taking the $\max(\cdot)$ of both ensures the distance is symmetric $d_H(\boldsymbol{P}_{GT}, \boldsymbol{P}_R) = d_H(\boldsymbol{P}_R, \boldsymbol{P}_{GT})$. This metric was computed using functionality from the Scipy Python library. The lower the Hausdorff Distance, the better the reconstruction performance.

*c. Mesh Similarity*

The Mesh Similarity measured the accuracy of normals from the reconstructed mesh $\mathcal{M}$. It was defined as the mean dot product between the normals of points in $P_{GT}$, which were treated as references, and the normals of the nearest faces of $\mathcal{M}$. This metric was computed by Equation (26)

$$\text{Mesh.sim}(\boldsymbol{P}_{GT}, \mathcal{M}) = \frac{1}{|\boldsymbol{P}_{GT}|} \sum_{x_i \in P_{GT}} \max(\boldsymbol{n}_{F_i} \cdot \boldsymbol{n}_{x_i}, -\boldsymbol{n}_{F_i} \cdot \boldsymbol{n}_{x_i}) \qquad (26)$$

where $F_i$ is the nearest face of $\mathcal{M}$ to point $x_i \in \boldsymbol{P}_{GT}$, and each $\boldsymbol{n} \in \mathbb{R}^3$ is a unit normal vector. The nearest faces were efficiently computed using functionality from the Trimesh Python library [70]. Further, recall that $\mathcal{M}$ was extracted from the generated SDF using the Marching Cubes algorithm (see Figure 16). This algorithm does not guarantee oriented normals [48] on $\mathcal{M}$ faces. To allow for misorientations due to Marching Cubes, the $\max(\cdot)$ dot product using the normal generated from Marching Cubes $\boldsymbol{n}_{F_i}$ and its flipped counterpart $-\boldsymbol{n}_{F_i}$ was computed prior to the summation, as seen in Equation (26). The mesh similarity metric ranged from 0 to 1, where 1 meant perfectly matching normals.

### 4.4.2 Performance evaluation

The aforementioned metrics were computed across all geometries in the training and testing datasets, and for all Auto-Decoder configurations. This computation resulted in metric values for each geometry and these values were collected as distributions and are visulised the violin plots in Figure 17.

Several noteworthy conclusions can be drawn from Figure 17. Excellent agreement between performance on training and testing datasets for all Auto-Decoder configurations can be seen when comparing Figures (a) and (b). This agreement provides evidence for the inexistence of overfitting and suggests the Auto-Decoders did indeed correctly learn the geometry space. No clear gain in performances can be seen between Auto-Decoders trained using the explicit learning approach (Explicit Nets 1-4). Although, Explicit Net 2 performed marginally better in terms of Chamfer Distance but was on par with its counterparts for the other two metrics. On the other hand, Implicit Net performed significantly better than all Explicit Nets across all metrics. This vast improvement in performance suggests the implicit learning approach was highly effective at learning stamping geometries when compared to the explicit learning approach. This result is further unpacked in the following.

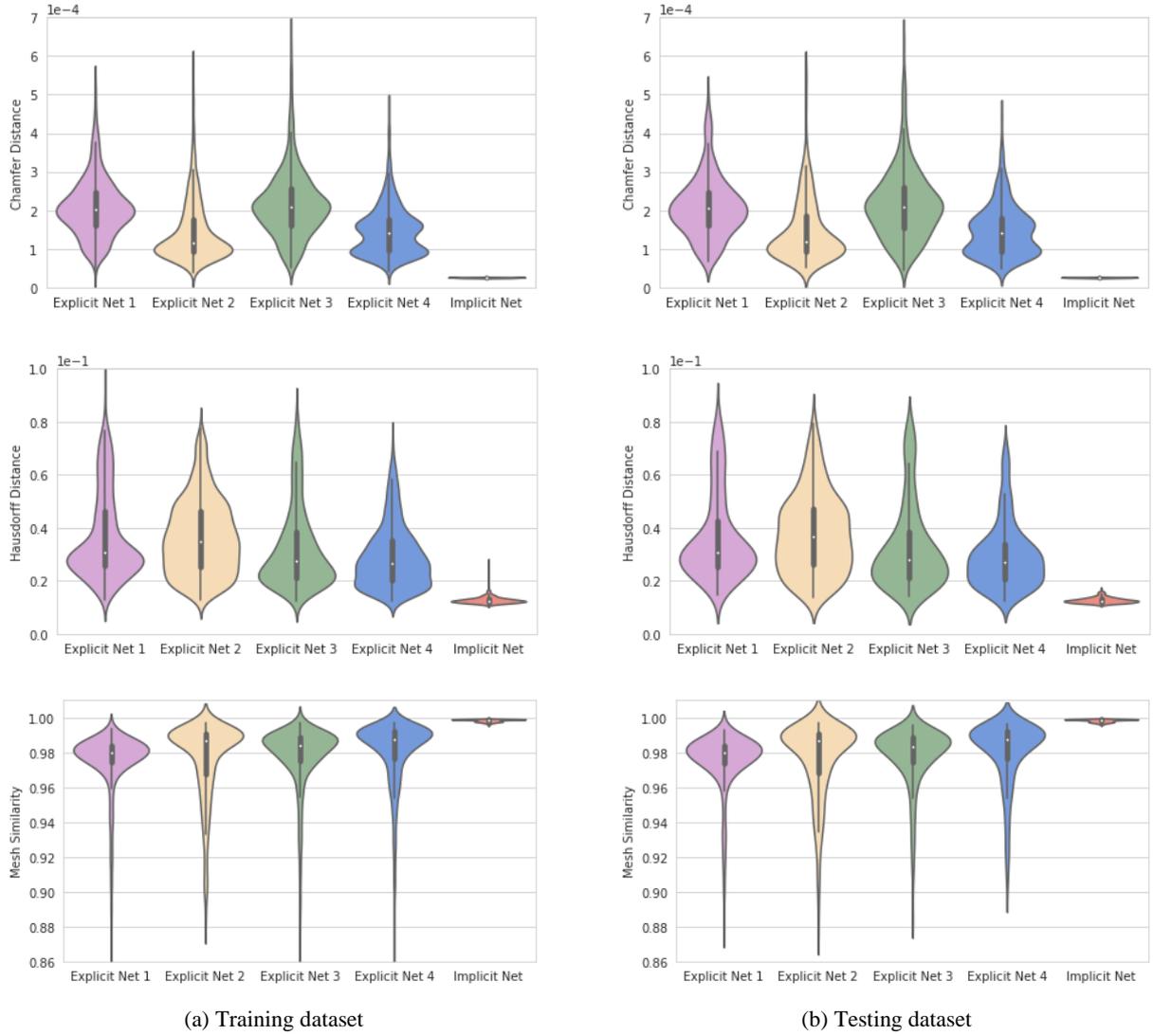

(a) Training dataset          (b) Testing dataset

Figure 17 Distributions of Chamfer Distance (top row), Hausdorff Distance (middle row) and Mesh Similarity (bottom row) for networks trained using the explicit learning approach (Explicit Nets 1-4) and implicit learning approach (Implicit Net).

Figure 18 quantifies the fraction of surface coverage that meets performance metrics $\rho$ between reconstructions and ground truths across all geometry samples in the testing dataset. In Figure (a), $\rho$ represents unsummed Chamfer Distance and is plotted on the X-axis. Plotted on the Y-axis is the fraction of points $x \in P_R$ that satisfy $\overrightarrow{d_C}(x, P_{GT}) \leq \rho$, where $\overrightarrow{d_C}(x, P_{GT})$ is the one sided Chamfer Distance from a point $x \in P_R$ to $P_{GT}$, i.e., the elements being summed in the second term of Equation (24). In Figure (b) $\rho$ represents unsummed Normals Similarity and is plotted on the X-axis. Plotted on the Y-axis is the fraction of faces $F \in \mathcal{M}$ that satisfy $\max(n_F \cdot n_x, -n_F \cdot n_x) \geq \rho$ and here $x$ is the closest point in $P_{GT}$ to face $F$.

Once again, Implicit Net significantly outperformed all Explicit Net variants in both performance metrics. The plots in Figure 18 show that Implicit Net had higher mean values and tighter standard deviation bands across all X-axis values. These outcomes suggest consistently high performance across all unseen geometries in the testing dataset. Results at the black dashed and dotted lines in Figure 18 are reported in Table 3 for further quantitative evaluation.

Table 3 Mean values ± standard deviations of selected values from Figure 18

| Auto-Decoder name | Chamfer Distance ≤ 1.5e-4 (Black dashed line in Figure 18(a)) | Chamfer Distance ≤ 2.5e-4 (Black dotted line in Figure 18(a)) | Normals Similarity ≥ 0.9 (Black dotted line in Figure 18(b)) |
|---|---|---|---|
| Explicit Net 1 | 0.19 ± 0.19 | 0.69 ± 0.20 | 0.93 ± 0.03 |
| Explicit Net 2 | 0.40 ± 0.16 | 0.84 ± 0.07 | 0.93 ± 0.05 |
| Explicit Net 3 | 0.11 ± 0.16 | 0.44 ± 0.32 | 0.94 ± 0.03 |
| Explicit Net 4 | 0.25 ± 0.18 | 0.74 ± 0.21 | 0.95 ± 0.04 |
| Implicit Net | 0.85 ± 0.03 | 1.00 ± 0.02 | 1.00 ± 0.01 |

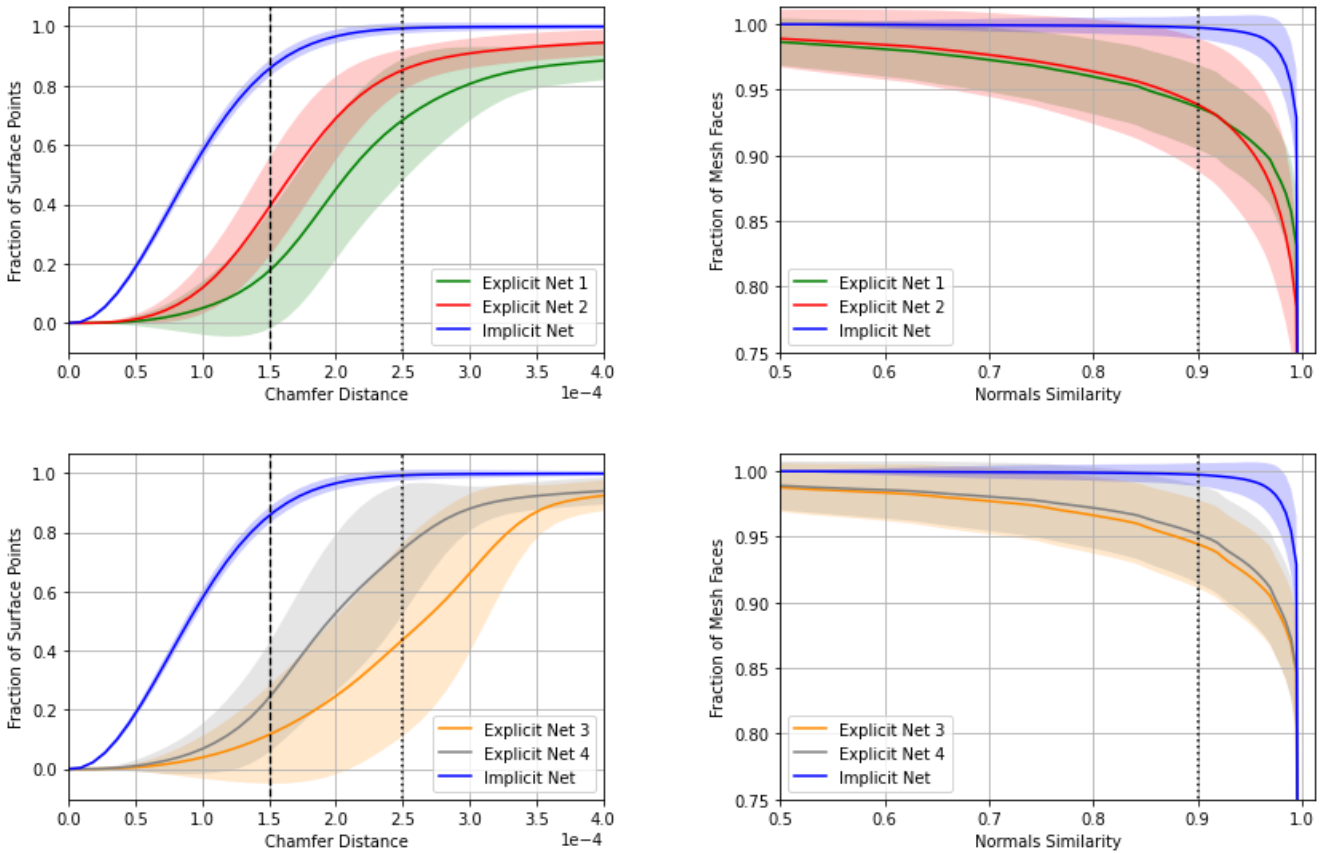

(a) One sided Chamfer Distance from reconstruction to ground truth  (b) Similarity of normals between reconstruction and ground truth

Figure 18 Testing dataset surface coverage versus performance metrics for various Auto-Decoder networks: Explicit Nets 1-2 and Implicit Net (top row), and Explicit Nets 3-4 and Implicit Net (bottom row). Solid curves represent means and shaded bands represent standard deviations across all samples in the testing dataset.

The excellent performance of Implicit Net on the testing dataset and across all metrics suggests that this network could generate realistic geometries during optimisation. A qualitative evaluation was next performed to further investigate this suggestion.

### 4.5 Qualitative evaluation of shape representation performance

To interpret the results of the quantitative evaluation, a qualitative evaluation was performed by visualising reconstructed surface meshes $\mathcal{M}$ obtained from the trained Auto-Decoder networks. Figure 20 compares surface meshes $\mathcal{M}$ reconstructed from Explicit Net 1 and Implicit Net with points sampled from the ground truth CAD surfaces $\boldsymbol{P}_{GT}$. The figure shows that although the surfaces from Explicit Net were mostly in agreement with the ground truth points, there were differences in local areas, such as radii. Some of these differences are indicated by the arrows in Figure (a). On the other hand, excellent qualitative correspondence was found between the surfaces from Implicit Net and the ground truth points, seen in Figure (b).

For further qualitative evaluation, Figure 20 compares 2D image projections of three surfaces which were reconstructed from Explicit Nets 1-2 and Implicit Net with 2D projections of their CAD ground truth counterparts. It was found that all the networks did indeed reconstruct geometries well on the global scale. The height values agreed between reconstructions and ground truths, and the reconstructions did replicate box corner geometries. However, this figure further confirms that reconstructions from Explicit Nets 1-2 failed to capture local radii features, as seen from their difference images. In contrast, near indistinguishable images are seen when comparing the reconstructions of Implicit Net with ground truths.

The improved performance using Implicit Net can be accredited to its adopted implicit learning approach (see Section 4.2.2) whereby the geometric properties of SDFs were learnt. In particular, the loss function contained a term that operated on the zero-level-set of the SDF directly (i.e., the surface geometry), as presented in Equation (19). In addition, the surface normal scaling factor in Equation (20) further promoted the accurate reconstruction of local geometric features on the surface. In contrast, the explicit learning approach was lacking such a surface loss and could only promote regression of SDF values close to the surface but not on it.

Recall from Figure 5 and Section 2.2 that manufacturability assessments were conducted during each optimisation iteration in the proposed optimisation platform. Further, Attar *et al.* [15,31] showed that local radii features are critical to determining manufacturing performance through sheet stamping. It is these local features that often require optimising to meet manufacturing constraints [14,15] and thus demand high quality representations. Therefore, since the Explicit Nets failed to reconstruct these critical features, the explicit approach was deemed unsuitable for representing sheet stamping geometries and therefore unsuitable for optimisation. Since Implicit Net provided high quality reconstructions of local features, it was deemed suitable for optimisation and was used hereafter as $f_{\theta_1}$ in Figure 5.

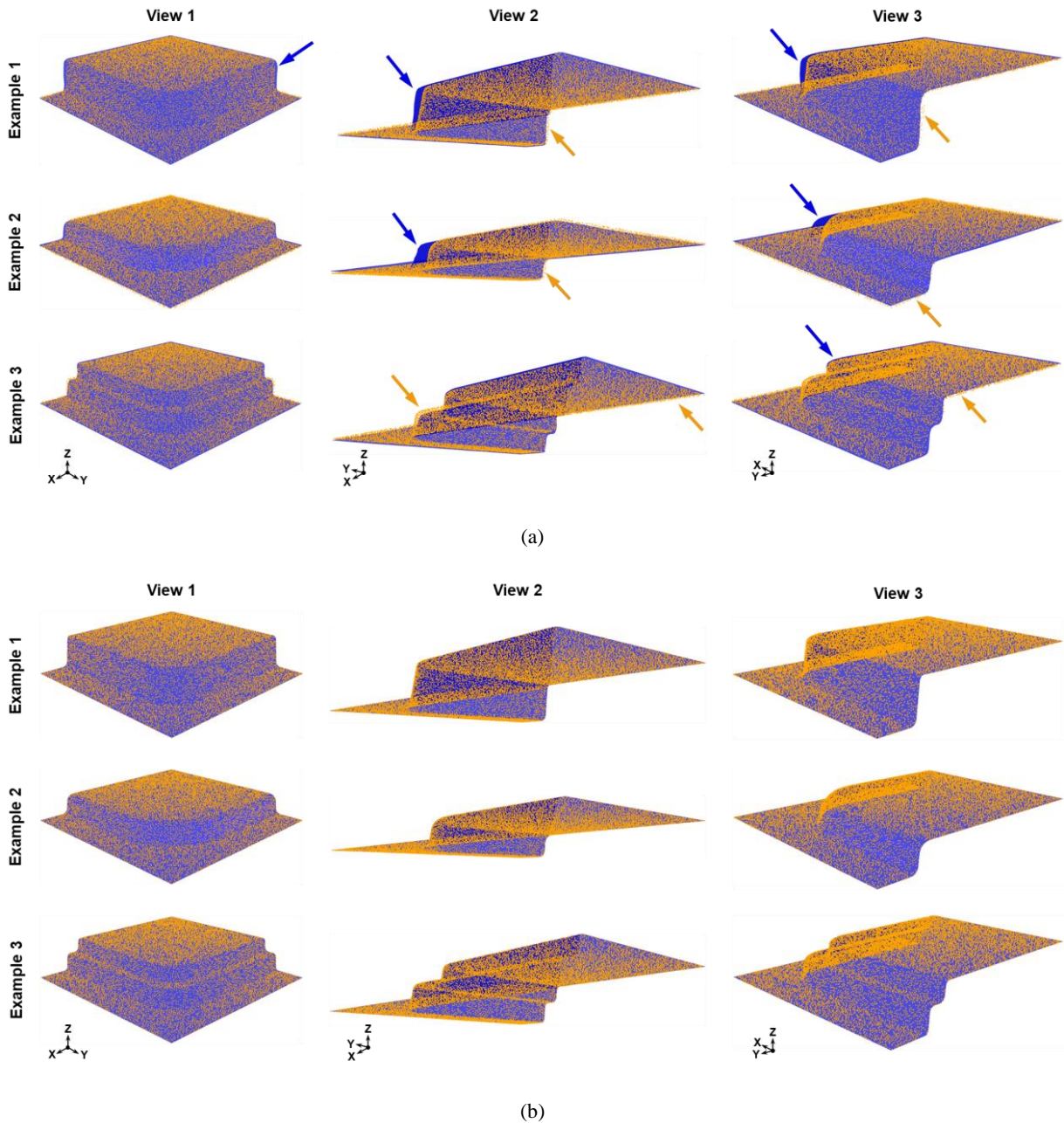

Figure 19 Comparison between reconstructed surface meshes $\mathcal{M}$ (purple surfaces) and ground truth surface points $\boldsymbol{P}_{GT}$ (orange points) for three representative geometries from the testing dataset. Reconstructions generated using (a) Explicit Net 1 and (b) Implicit Net. Arrows in (a) indicate zones of misalignment between $\mathcal{M}$ and $\boldsymbol{P}_{GT}$. All images taken with the perspective camera projection type for best figure clarity.

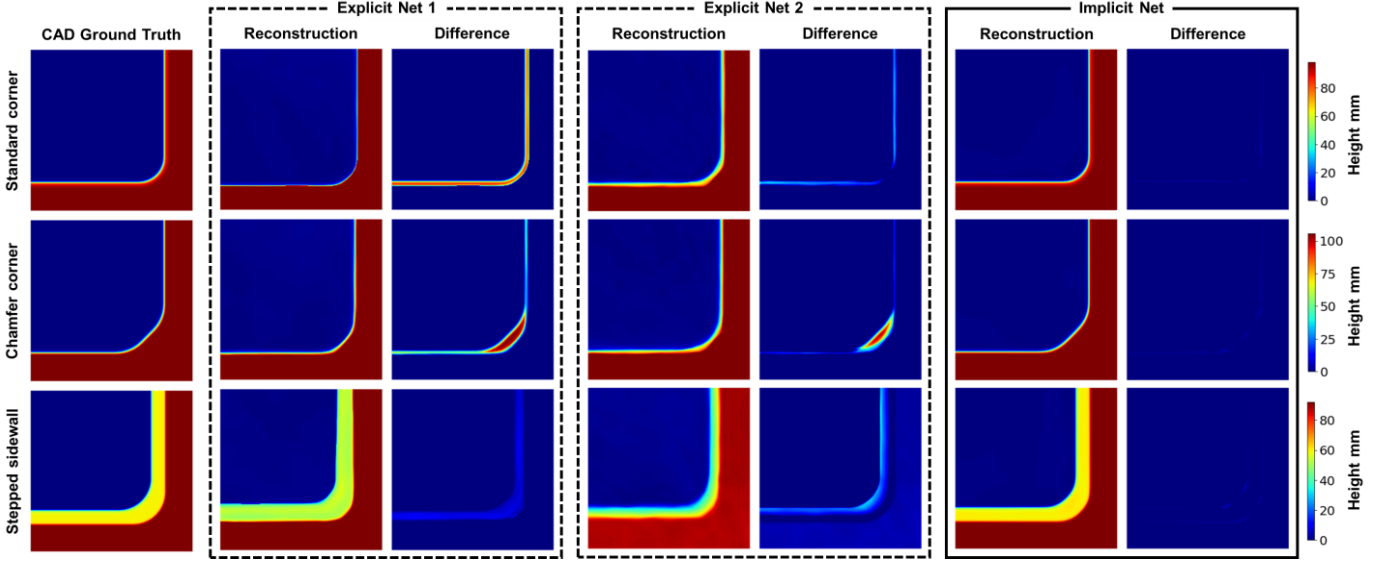

Figure 20 Comparison between 2D projected height maps of reconstructed surfaces from Explicit Nets 1-2 (dashed line rectangles) and Implicit Net (solid line rectangle) with CAD ground truths for three representative geometries from the testing dataset. Colourbar limits set equal across columns for comparison between reconstruction methods.

## 4.6 Continuity of the learnt geometric latent space

The latent space refers to the learnt low dimensional space that contained the inferred latent vectors $\boldsymbol{z} \in \mathbb{R}^{128}$ for the considered stamping geometries. The term *low dimensionality* here is used in relation to high dimensional data representations, such as images or meshes. In the case of an image, the dimensionality is equal to the total number of pixels (e.g., $256 \times 256 = 65{,}536$ pixels). In the case of a mesh, the dimensionality is equal to the total degrees of freedom of mesh vertices (e.g., $10{,}000$ vertices $\times$ 3 orthogonal directions $= 30{,}000$ degrees of freedom).

In this subsection, the continuity of the latent space learnt by Implicit Net is presented. As mentioned by Wang *et al.* [32], when continuity is combined with low dimensionality, different vectorised directions in the latent space encode geometrically meaningful embeddings. These embeddings were leveraged for the proposed optimisation platform.

### 4.6.1 Organisation of the latent space

The learnt latent space was found to be organised according to the natural similarity between different geometries. To visualise this organised space, three-dimensional principal component analysis (PCA) was performed on latent vectors inferred from geometries in the training and testing datasets. Following PCA, the principal components of the latent vectors were obtained, and this reduced the space from $\mathbb{R}^{128}$ to $\mathbb{R}^3$. These principal components are plotted in Figure 21, where it was found that geometries from the same parameterisation scheme cluster together in the latent space. This clustering can be interpreted as the latent space being organised in terms of global geometric features. A mild overlap was found between red and green points, which correspond to geometries that were similar to both standard and chamfer corners (e.g., chamfer corners with small chamfer lengths). In contrast, the blue points, which corresponded to stepped sidewalls, were further away because their stepped feature made them dissimilar to the other geometries.

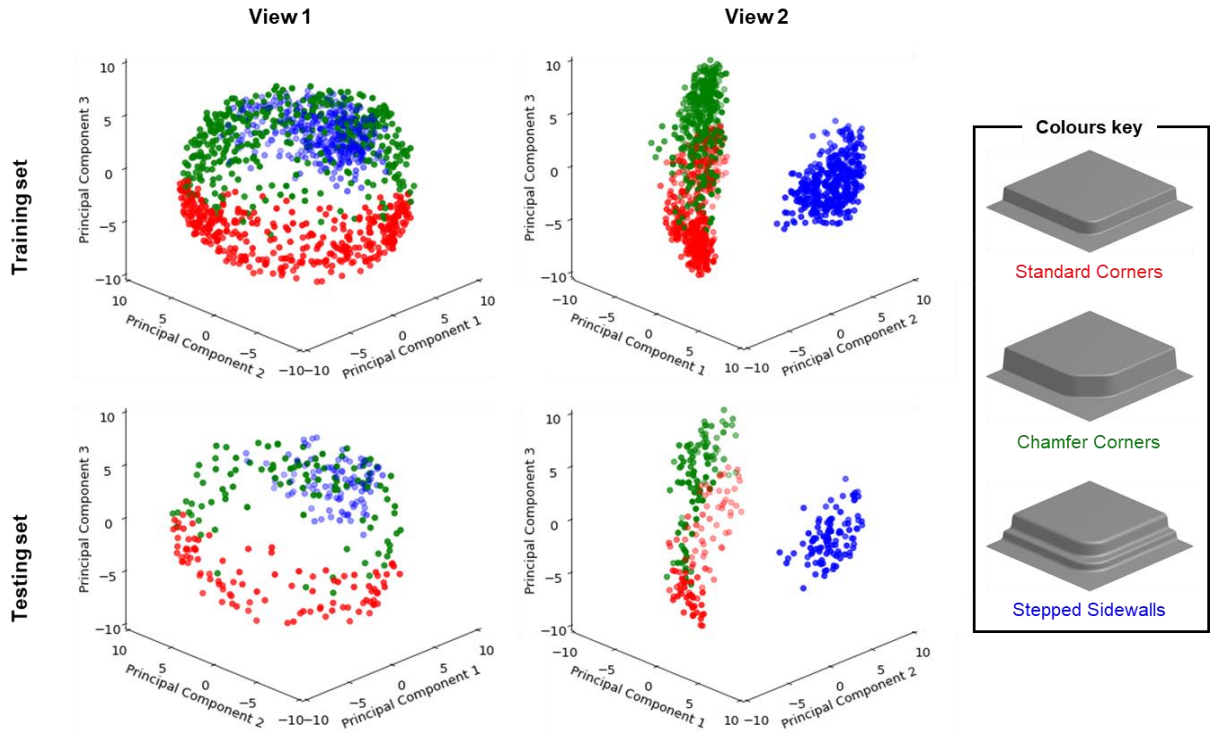

Figure 21 Three dimensional PCA of the learnt latent space. Coloured clusters denote points belonging to different geometry subclasses, and therefore global geometric features, and are labelled in the colours key. Views selected for best figure clarity.

To further show the organisation of the latent space, the three coloured clusters in Figure 21 are plotted separately in Figure 22. In each plot, selected local geometric features (i.e., height, radius, or similar) defined in Figure 6(a) are superimposed as colours. The smooth transitions between different colours show that the latent space was further organised in terms of these local geometric features in addition to global features.

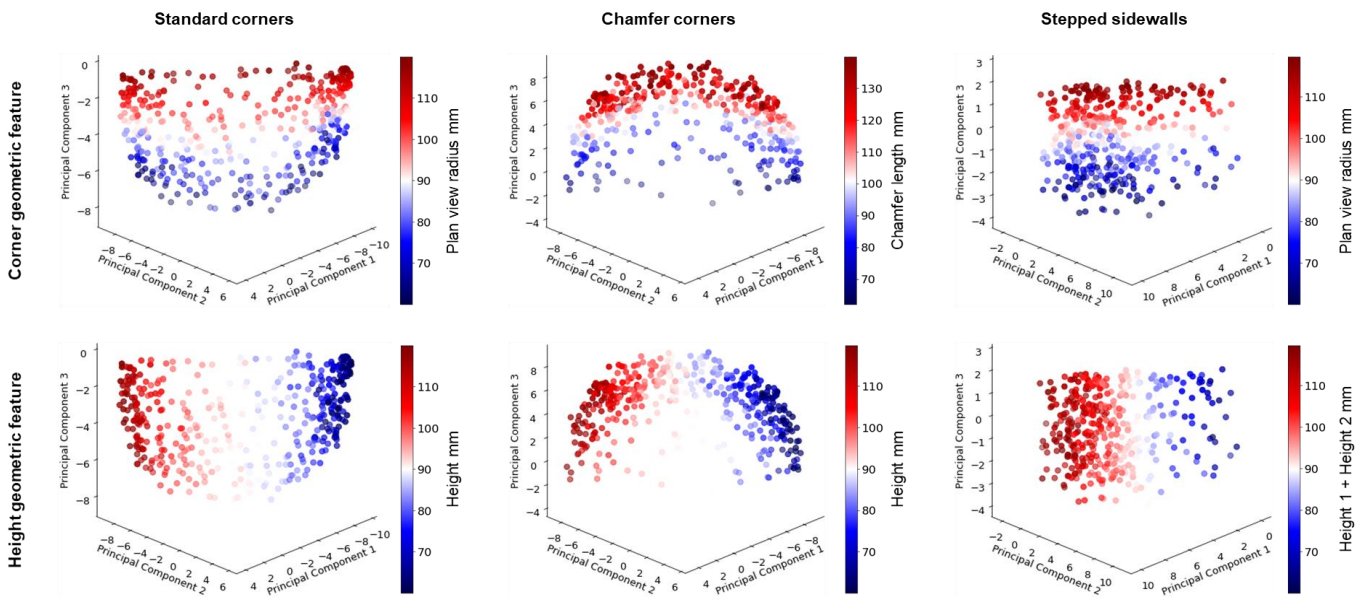

Figure 22 Three dimensional PCA of the learnt latent space where each cluster in Figure 21 is plotted here individually. The colour bars represent values of a local geometric feature. The geometric features follow those defined in Figure 6(a). Views selected for best figure clarity.

In summary, it was found that the latent space encoded meaningful and well organised geometric information on a global and local scale. Consequently, the proximity of two points in the latent space provided a measure of similarity between two geometries represented by these points. This distance metric was exploited during optimisation and further explained in Section 6.

### 4.6.2 Interpolation in the latent space

To evaluate the learnt latent space further, geometries generated by interpolating between latent vectors are presented in Figure 23. In Figure (a) the geometries shown in blue were generated from latent vectors that were averaged from those of the adjacent grey test set geometries. It is evident that these generated geometries naturally combine common geometric features of their adjacent test set constituents. In Figure (b), the geometries shown in blue were generated from latent vectors that were linear combinations of those of the two grey test set geometries, $\mathbf{z}_A$ and $\mathbf{z}_B$, according to Equation (27).

$$\mathbf{z}_\alpha = (1-\alpha)\mathbf{z}_A + \alpha \mathbf{z}_B \tag{27}$$

Here, $\mathbf{z}_\alpha$ is the latent vector at the value of the scalar $\alpha \in [0,1]$. The smooth changes between the geometries in Figure (b) demonstrate that robust free-morphing of sheet stamping geometries was possible by exploring the learnt latent space. As a separate note, notice the well captured sharp fillet radii in all geometries shown.

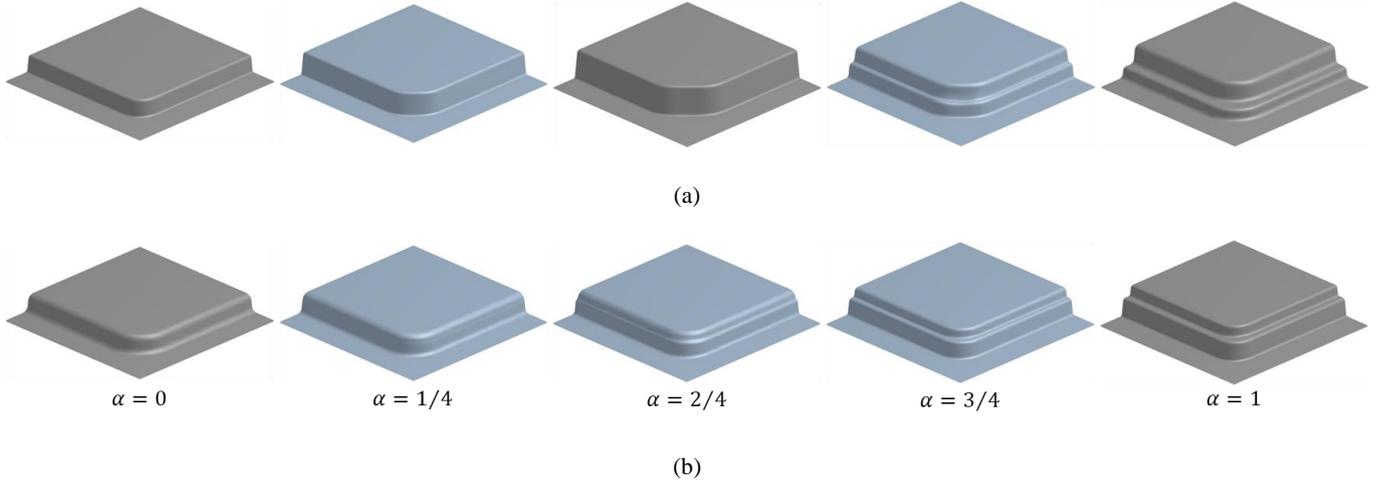

(a)

$\alpha = 0$   $\alpha = 1/4$   $\alpha = 2/4$   $\alpha = 3/4$   $\alpha = 1$

(b)

Figure 23 Reconstructions of test set samples (in grey) and generated geometries (in blue). In (a), the generated geometries were decoded from latent vectors that were averaged from those of adjacent grey geometries. In (b), the generated geometries were decoded from latent vectors that were obtained by linearly interpolating between those of the grey geometries.

## 5 Development of the manufacturability assessment surrogate model

The development of the manufacturability assessment surrogate model is outlined in this section. Details given include steps to acquire training data, brief explanation of the model architecture and a performance evaluation. This model is needed to rapidly and differentiability incorporate manufacturability information into the optimisation platform.

### 5.1 Manufacturing process simulations

Training data for the manufacturing performance evaluator $g_{\theta_2}$ was sourced from FE forming simulations. The selected manufacturing process was the HFQ process [5]. Details on the HFQ simulation setup and forming characteristics of the three corner subclasses introduced above are now given.

#### 5.1.1 HFQ simulation setup

The CAD geometries generated from the DoE described in Section 3.2 were converted into FE meshes for the forming simulations. To automate the meshing of these CAD geometries, the TCL programming language was used within the commercial FE pre-processing software HyperMesh. The FE software PAM-STAMP was utilised for the HFQ stamping simulations, due to its specialisation for sheet stamping. PAM-STAMP's scripting interface, together with the Python programming language, was used to automate the loading of the prepared mesh files and launch the computations. To simulate non-isothermal HFQ conditions, coupled thermo-mechanical simulations were conducted. The remainder of the simulation set up can be found in Section 3.2 and 3.3 of the original paper which proposes CNN based surrogate models for hot stamping simulations by Attar *et al.* [31]. In that paper, details of FE models, adaptive FE mesh refinement and material model, that was for AA6082 calibrated under HFQ conditions [71], were presented by the authors in detail. Since the main focus of this study was optimisation of component geometries, the processing parameters were assumed constant, and are given in Table 4.

Table 4 Main process and simulation parameters

| Parameter | Value |
| --- | --- |
| Initial workpiece temperature °C | 500 |
| Initial tooling temperature °C | 25 |
| Stamping speed mm/s | 500 |
| Spacer thickness mm | 2.2 |
| Friction coefficient (all interfaces) | 0.1 [72] |
| Velocity scaling factor | 10× |

### 5.1.2 Representative HFQ simulation results

To better understand the stamping characteristics of geometries from the three geometry subclasses outlined in Section 3.1, and ensure interpretable optimisation results, representative thinning fields are plotted in Figure 24. The thinning fields were used here to define the manufacturing performance. In Figure 24, the geometric parameters were kept approximately consistent between the three parameterisation schema to allow for comparison. It was found that stamping a standard corner resulted in three noteworthy locations, as denoted by zones A1-A3. When forming a chamfer corner, zone A1 transitioned into two locally thinning zones denoted by B1 and the sidewall thinning B2 was more evenly spread when compared to zone A2. The thickening at zone A3 was also reduced and can be seen by B3. It was found that the stepped sidewall generally resulted in the most severe thinning cases. In particular, concentrated thinning was seen in zones C1-C3 for a range of stepped sidewall geometries. The thickening at zone C4 was found to be similar to that of zone A3 for the standard corner.

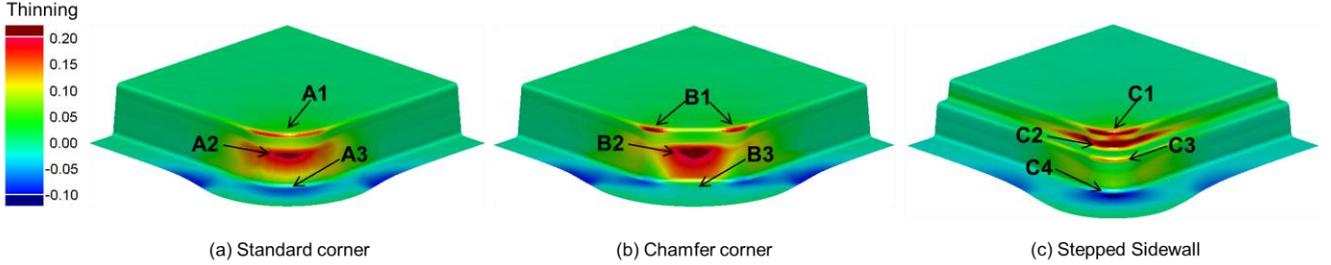

Figure 24 Representative thinning fields obtained from forming geometries from the three considered geometry subclasses under HFQ conditions.

It was found that adopting a Latin Hypercube DoE for geometries resulted in an even distribution of thinning field variants within the output target space. More formally, this meant that a good variation in thinning magnitudes at all of the highlighted zones shown in Figure 24 were obtained. Furthermore, the magnitudes of thinning at each location was dependant on the combination design parameter values for each geometry subclass. Based on these dependencies, for standard corners, Attar *et al.* [15] found that the location of maximum thinning switches between zone A1 and A2, depending on the combination of radii parameters. Therefore, by training $g_{\theta_2}$ in this work on this dataset would allow the network to predict a range of thinning fields and capture manufacturability trends when presented with unseen geometries (see Section 5.4). Consequently, this predictive capability would enable the network to incorporate manufacturing constraints (e.g., a constraint on maximum thinning value which is robust to a location change between zone A1 and A2 [15]), and use these to drive the optimisation process and thus enable non-trivial geometry changes to be made.

### 5.2 Training data pre-treatment

Recall from Figure 4 that 2D orthographic projections of 3D stamping die surfaces were performed and these were possible without loss of spatial information. Further recall from Figure 5 that during each optimisation iteration, these 3D stamping geometries were generated from $f_{\theta_1}$ and not from CAD ground truths. Figure 25 highlights the pixel-wise differences between 2D projections from CAD ground truths and from $f_{\theta_1}$ reconstructions (Implicit Net used). The colourbars show that although these differences were over an order of magnitude smaller than absolute height values, they were not zeros. Therefore, the input images for training and testing the manufacturing performance evaluator $g_{\theta_2}$ were 2D projections of reconstructions from $f_{\theta_1}$ and not from CAD ground truths. It was found that this step was needed to avoid these small pixel-wise differences being interpreted as noise by $g_{\theta_2}$ which would otherwise interfere with its performance. To perform the actual 2D projections, the rasteriser from the Pytorch3D library was used [59]. The resulting 2D projections had a resolution of 256 × 256 to capture small geometric features such as sharp tool radii. These images of die geometries formed the inputs to $g_{\theta_2}$ for its training and testing datasets.

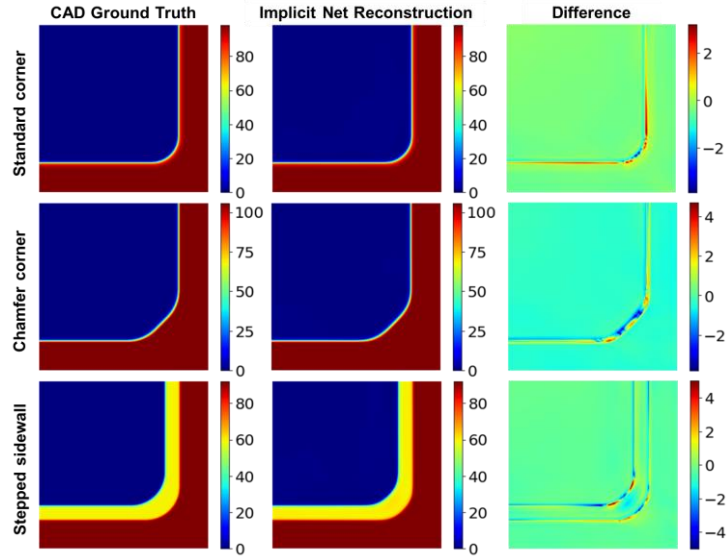

Figure 25 Comparison between 2D projected height maps of reconstructed surfaces from Implicit Net with CAD ground truths for three representative geometries from the testing dataset. Colourbars set to show small differences between ground truth and reconstruction images. Colourbar height values in mm.

The target images were also projected onto 2D images from 3D FE simulation results like those shown in Figure 24. The full process for the 2D projection of these images was previously detailed by the authors in [31]. Briefly, the deformed 3D simulation mesh was first undeformed back into the 2D blank position by subtracting the post-stamping nodal displacements from the deformed 3D nodal positions. The result was treated as a 2D point cloud which had a thinning strain value at each node. These thinning strain values were then linearly interpolated onto 256 × 256 cartesian grids and the result formed the target images for the training and testing datasets of $g_{\theta_2}$.

Once the input and target images (i.e., height maps and thinning field maps respectively) were prepared, the final aspect to the data pre-treatment was the ordering of the dataset. The input-target image pairs were ordered as previously outlined in Section 4.3.2, and this ordering is illustrated in Figure 26.

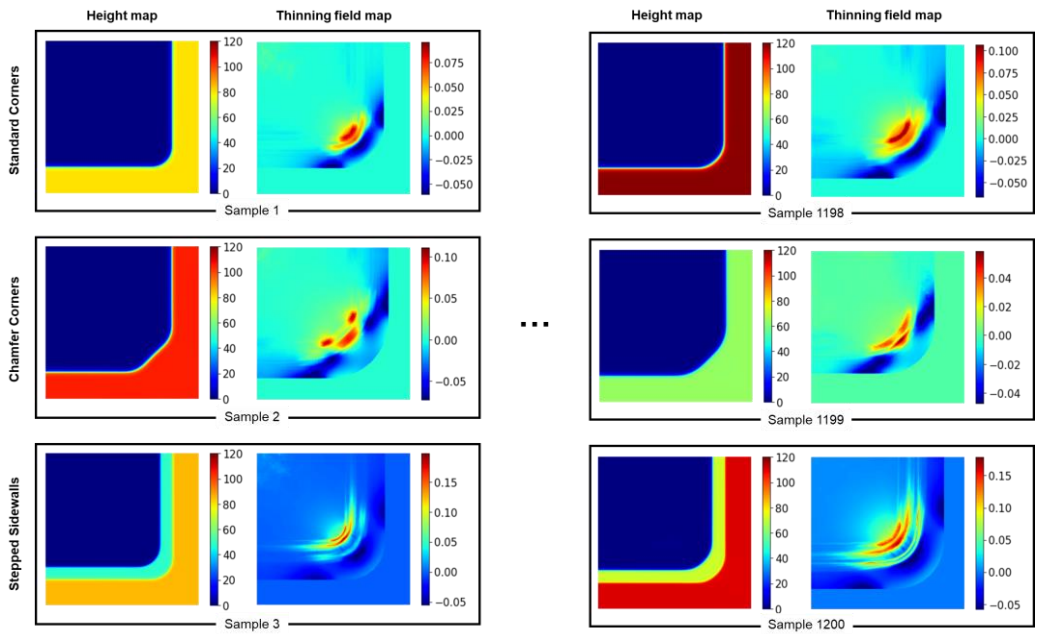

Figure 26 Ordering of samples in the training dataset for the manufacturability surrogate model; random geometries from all 3 geometry subclasses were arranged periodically.

## 5.3 Model architecture and training

The Res-SE-U-Net architecture presented in Attar *et al.* [31] was employed as the manufacturing performance evaluator $g_{\theta_2}$. The reader is referred to the original paper for full details of this architecture, and these details include network layers, activation functions and use of batch normalisation. Briefly, the model consisted of a down sampling encoder made up of convolutional layers, a bottleneck with Res-SE layers and an up sampling decoder made up of up convolutional layers. Details on Res-SE layers can be found in [31,73,74].

A schematic of this encoder-decoder architecture was previously presented in Figure 4(b). The role of the encoder was to identify characteristic geometric features within a given height map through a series of convolutional layers and store them in a spatially compact (in terms of image resolution) bottleneck. Examples of such design features include corner radii and total geometry height which were detected from the height maps. The decoder was responsible for taking this compact information and up-scaling it through a series of up convolutional layers into an image of the thinning field. Skip connections were used to bridge the encoder and the decoder to reduce information loss during the encoding phase.

$g_{\theta_2}$ was trained on the dataset of height maps and corresponding thinning fields shown in Figure 26. The training batch size was selected to be 21. It was important to select a batch size that was a multiple of 3 as outlined in Section 4.3.2. The Adam optimiser [62] with default beta parameters of $\beta_1 = 0.9$ and $\beta_2 = 0.999$ were used. The learning rate was $5 \times 10^{-4}$ and was constant throughout training. Through the iterative optimisation process, the optimiser sought to find the combination of network parameters $\theta_2$ such that the mean squared error (MSE) between ground truth (target thinning field images) and network predictions was minimised.

## 5.4 Manufacturability prediction results

After training $g_{\theta_2}$, its performance was evaluated by comparing its thinning field predictions with those of ground truths based on FE simulation results from the unseen testing set. A comparison of thinning distributions from FE ground truths (GT) and network predictions (PD) for five random test set cases are presented in Figure 27. The maximum absolute error (MAE), defined as $|\max(PD) - \max(GT)|$, is also reported in the figure for each case. For all cases, indistinguishable thinning distributions between network predictions and ground truths can be seen.

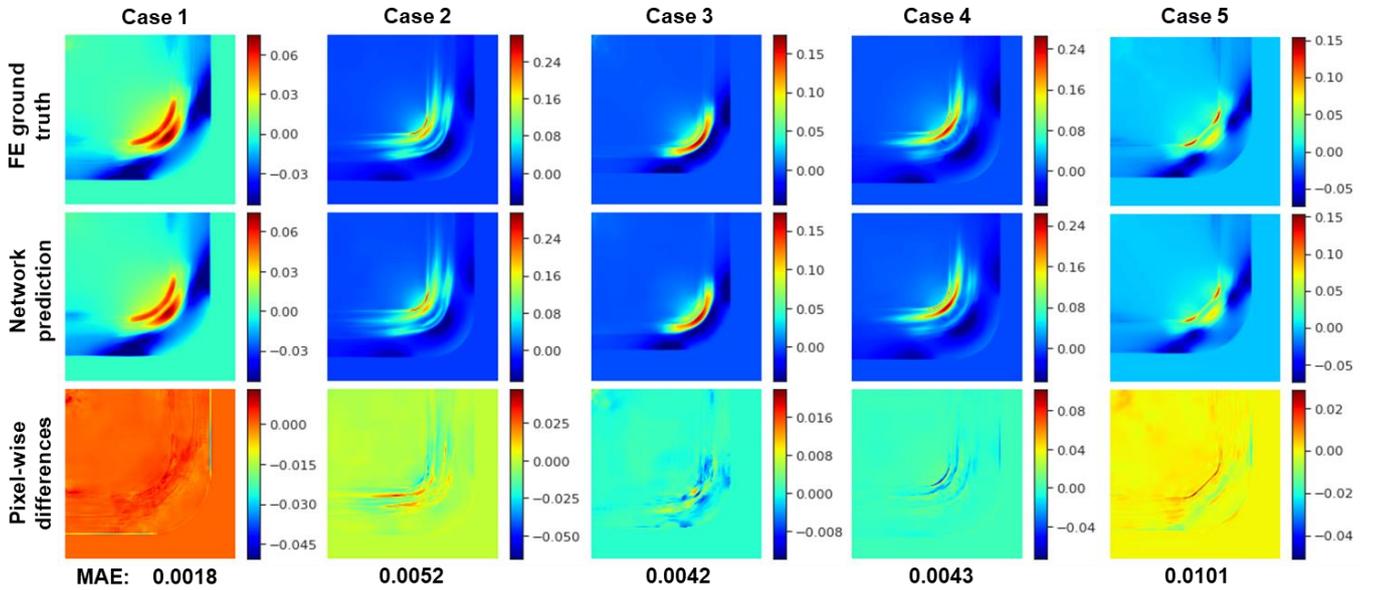

Figure 27 Comparison of thinning distributions from FE ground truths and network predictions for five random test set cases.

To further evaluate the overall training performance of $g_{\theta_2}$, the data distributions are visulised by the violin plots in Figure 28. In the figure headings, the Kullback-Leibler Divergences (KLDs) are reported, and these qualified the similarity between GT and PD distributions, where a value of 0 suggests identical distributions. The max and mean absolute thinning values of each image in the training and test sets were calculated for both GT and PD and these values were used to represent the data distributions in the figure. Excellent agreement can be seen between GT and PD distributions and this agreement is also confirmed by the small values of KLDs.

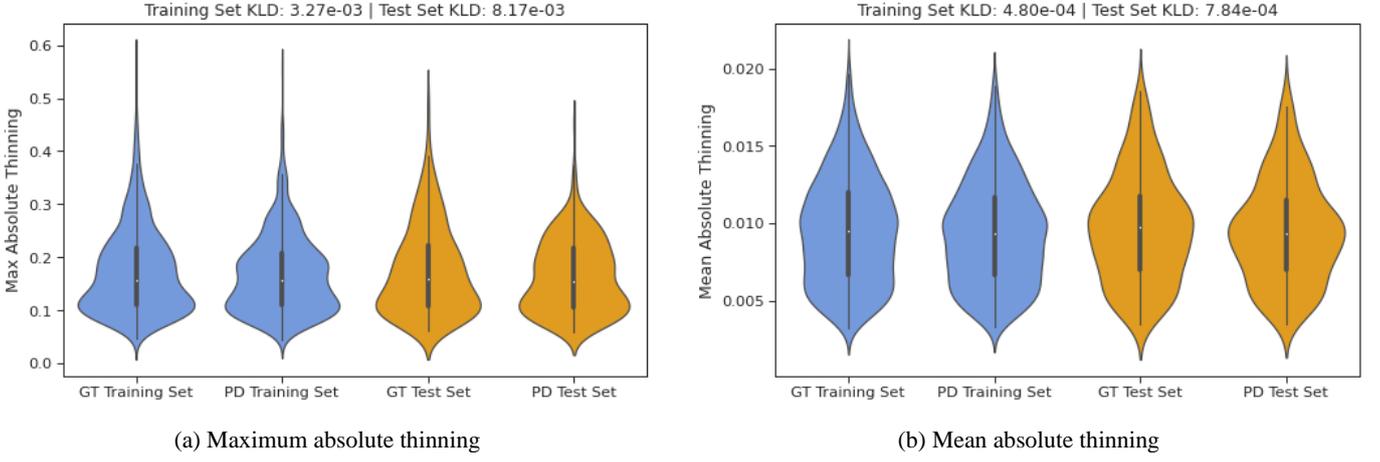

(a) Maximum absolute thinning  (b) Mean absolute thinning

Figure 28 Comparison of dataset distributions and KLD between FE ground truths (GT) and network predictions (PD) for testing and training sets.

Since $g_{\theta_2}$ did not solve any system dynamics iteratively like an FE solver would, it learnt the physics of the HFQ forming process from data supplied during training. Following the evaluation of the presented results, the performance of $g_{\theta_2}$ in predicting the manufacturability of unseen geometries was deemed sufficiently acceptable. Therefore, this network was employed as the manufacturing performance evaluator in the proposed optimisation platform.

# 6 Design optimisation

The optimisation platform was set up by following the developments presented in the former sections of this paper. This section presents optimisation results obtained from the platform for two different tasks. Optimisation was performed with respect to the latent vector $\boldsymbol{z}$ and starting with an initial latent vector that best represents the initial geometry to be optimised. To obtain this initial latent vector, decoder inference was first performed to minimise Equation (23) and a schematic of this procedure was previously presented in Figure 13. Figure 29 shows the convergence history of the terms in Equation (23) for an arbitrary geometry. This step was part of the pre-optimisation phase which was completed before each optimisation process and was further outlined in Section 2.3.1.

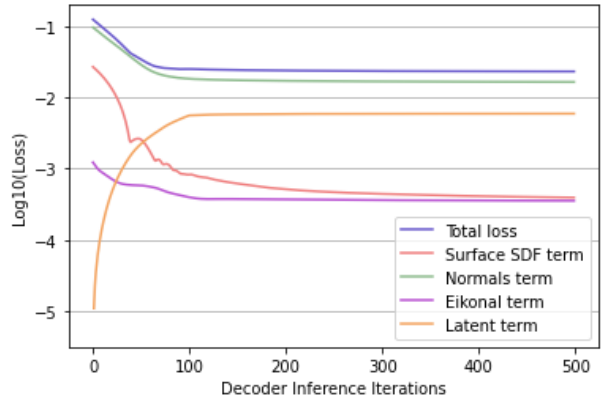

Figure 29 An example plot of the convergence histories of the terms in Equation (23) for decoder inference. At convergence, a latent vector that best describes the initial geometry was obtained.

## 6.1 Task 1: Differentiable changes of sheet stamping geometry

Task 1 was to search the geometry space to find the geometry that was responsible for a target manufacturing performance field (in this study, thinning field). The purpose of this task was to demonstrate the end-to-end differentiability of the optimisation pipeline which was presented in Figure 2. The objective function that was to be minimised is presented in Equation (28)

$$\arg\min_{\boldsymbol{z}} \mathcal{L}_{task1}(\boldsymbol{z}) = \arg\min_{\boldsymbol{z}} \|\boldsymbol{M}_T - \boldsymbol{M}(\boldsymbol{z})\|_2^2 \tag{28}$$

where

$$M(z) = g_{\theta_2}\left(\phi\left(f_{\theta_1}(z, X)\right)\right) \quad (29)$$

and $M_T$ is the 2D thinning field of the target geometry and was constant. $M(z)$ is the generated 2D thinning field of the geometry that was represented by the latent vector which was being optimised $z$ and this field was therefore evolving during optimisation. $X$ is the uniform grid that defined the SDF domain, $f_{\theta_1}$ is the Auto-Decoder which generated an SDF that was conditioned on $z$, $\phi$ is the combined marching cubes and differentiable rasteriser steps, and $g_{\theta_2}$ is the manufacturability surrogate model.

Computing $\mathcal{L}_{task1}(z)$ was the result of the forward pass which occurred once per optimisation iteration, as outlined in Section 2.3.2. Equation (28) was substituted into Equation (12) and the gradient $\partial \mathcal{L}_{task1}/\partial z$ was computed. Using this gradient, the Adam optimiser performed gradient-based updates of $z$ at each iteration to minimise Equation (28).

The history and result of this optimisation process is presented in Figure 30. Figure 30(a) shows that the objective function was smoothly minimised during optimisation and asymptotically tending towards zero. The evolution of the thinning fields can be seen in Figure 30(b) as $M(z)$ slowly tended towards $M_T$ and changed the geometry subclass to do so. The resulting geometries smoothly transitioned from the stepped sidewall on the left to a chamfer corner on the right that closely matched the target geometry. This result demonstrates that manufacturing performance information can be backpropagated in order to non-parametrically optimise the stamping geometry.

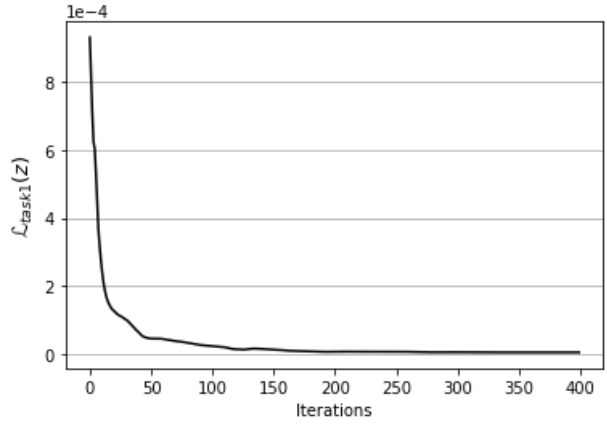

(a) History of $\mathcal{L}_{task1}(z)$ in Equation (28)

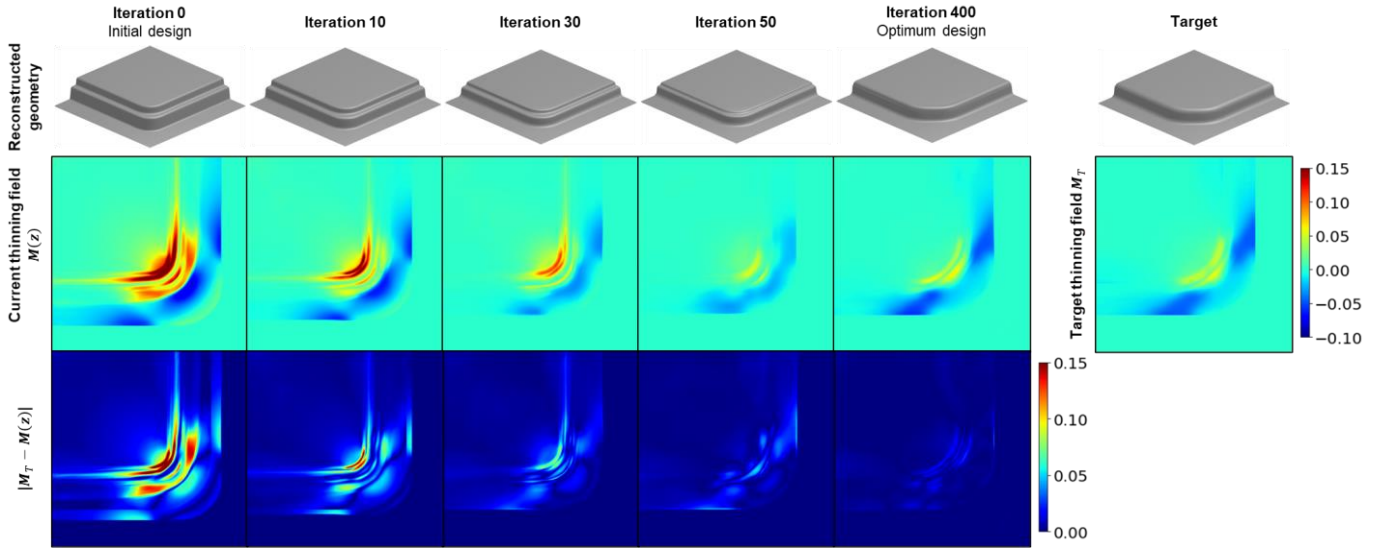

(b) Visualisation of reconstructed geometries and thinning fields during optimisation

Figure 30 Results for minimising the difference between thinning fields from an initial geometry and a target geometry with respect to the latent vector $z$. Performing this minimisation transformed the initial geometry to closely match the target geometry through their thinning field information only.

## 6.2 Task 2: Sheet stamping geometry design subject to manufacturing constraints

Task 2 was to optimise an initial geometry to be successfully formed through a hot stamping process. This task is driven by todays industrial needs as component designers would need to adapt existing designs to be manufacturable through the latest state-of-the-art stamping processes [15,75].

### 6.2.1 Optimisation problem formulation

To perform this optimisation task, a new objective function in the form of a Lagrangian equation is proposed and presented in Equation (30)

$$\arg\min_{z} \mathcal{L}_{task2}(z) = \arg\min_{z} \left( \mathcal{L}_{similarity}(z) + \mathcal{L}_{constraint}(M(z)) + \mathcal{L}_{latent}(z) \right) \tag{30}$$

where

$$\mathcal{L}_{similarity}(z) = \frac{\lambda_1}{|z|} \|z_0 - z\|_2^2 \tag{31}$$

$$\mathcal{L}_{constraint}(M(z)) = \lambda_2 \text{ReLU}(\max(M(z)) - t_{max}) \tag{32}$$

$$\mathcal{L}_{latent}(z) = \frac{\lambda_3}{|z|} \|z\|_2 \tag{33}$$

and all $\lambda$ terms are weighting constants. This equation incorporated a manufacturing constraint and was formulated such that minimising the total equation would improve the manufacturability. The terms in this equation are now explained.

The $\mathcal{L}_{similarity}(z)$ term ensured that the deviation between the geometry that was being optimised and the initial geometry was minimised, and thus their similarity was maximised. This similarity term was needed to prevent the optimiser from finding non-feasible engineering solutions, and ensure the optimum design was as close as possible to an initial design. For example, if a Battery Box geometry with tight corners is required in practice, the optimiser may otherwise converge to a straight U-Channel geometry. The U-Channel is almost always easier to manufacture but would not satisfy the engineering needs. Formulating $\mathcal{L}_{similarity}(z)$ in terms of latent vectors took advantage of the continuous and well organised latent space which was presented and explained in Section 4.6.1. Here, the initial geometry was encoded in $z_0$ and the geometry being optimised was encoded in $z$. The strength of $\mathcal{L}_{similarity}(z)$ was tuned through the constant $\lambda_1$.

The $\mathcal{L}_{constraint}(M(z))$ term was the main driver for the optimisation and this term encapsulated the manufacturability information for the geometry encoded in $z$. Shown in Equation (29), $M(z)$ is the generated thinning field of the geometry that was encoded in $z$. $M(z)$ was computed during the forward pass which occurred once per optimisation iteration. Since $M(z)$ contained the entire post-stamping thinning field, it was used to define manufacturing constraints. In this paper, a constraint on the maximum thinning value was imposed and this formulation can be seen in Equation (32). Here, $t_{max}$ was a scalar that represented this maximum thinning constraint. The function $\text{ReLU}(x) := \max(0, x)$ was used to make this constraint inactive when $\max(M(z))$ did not exceed $t_{max}$ and active otherwise. When the constraint was active, $\mathcal{L}_{constraint}(M(z))$ imposed a large penalty on the Lagrangian in Equation (30). The strength of this penalty was tuned through the constant $\lambda_2$.

The $\mathcal{L}_{latent}(z)$ term was a standard regulariser on the latent vector. This term prevented the optimiser from finding latent vectors with magnitudes that were too high and thus would have led to the generation of unrealistic shapes. The strength of $\mathcal{L}_{latent}(z)$ was tuned through the constant $\lambda_3$.

### 6.2.2 Reformulating the objective function to facilitate differentiability

Recall from Section 2.3.3 that $\partial \mathcal{L}_{task}/\partial z$ was required in order to use manufacturability information to drive geometry updates and that this gradient had to be calculated using Equation (12). This requirement meant that Equation (30) needed to be differentiated with respect to $z$ to obtain $\partial \mathcal{L}_{task2}/\partial z$. To obtain this gradient, it was seen from Equation (9) that $\partial \mathcal{L}_{task2}/\partial M$ was required. However, the only term in Equation (30) that contained $M$ for this optimisation task was $\mathcal{L}_{constraint}(M(z))$ and this meant that the two terms struck through in Equation (34) would be zeros if $\partial\mathcal{L}_{task2}/\partial M$ were to be computed in this way.

$$\frac{\partial\mathcal{L}_{task2}}{\partial M} = \cancel{\frac{\partial\mathcal{L}_{similarity}(z)}{\partial M}} + \frac{\partial\mathcal{L}_{constraint}(M(z))}{\partial M} + \cancel{\frac{\partial\mathcal{L}_{latent}(z)}{\partial M}} \qquad (34)$$

Therefore, computing $\partial\mathcal{L}_{task2}/\partial z$ using Equation (9) and Equation (12) meant that the $\mathcal{L}_{similarity}(z)$ and $\mathcal{L}_{latent}(z)$ terms would become redundant. To overcome this potential redundancy, the Lagrangian in Equation (30) was recast to only contain the manufacturing constraint term, as seen in Equation (35).

$$\mathcal{L}_{task2}(z) = \mathcal{L}_{constraint}(M(z)) = \lambda_2 \text{ReLU}(\max(M(z)) - t_{max}) \qquad (35)$$

Then, the $\mathcal{L}_{similarity}(z)$ and $\mathcal{L}_{latent}(z)$ terms were inserted into Equation (12) but before the gradient with respect to $z$ was taken, as seen in Equation (36). Since these terms were not functions of $M$, this formulation was still valid. Further, since these terms were introduced into the formulation after $M$ was computed from Equation (35), the potential redundancy shown in Equation (34) was avoided. Using Equation (36), the Adam optimiser performed manufacturability-informed gradient-based updates of $z$ at each iteration to minimise Equation (30).

$$\frac{\partial\mathcal{L}_{task2}}{\partial z} = \frac{\partial}{\partial z}\left(\frac{1}{|V|}\sum_{v \in V} \alpha_v \frac{\partial\mathcal{L}_{constraint}(M(z))}{\partial f_{\theta_1}(z,v)} \cdot f_{\theta_1}(z,v) + \mathcal{L}_{similarity}(z) + \mathcal{L}_{latent}(z)\right) \qquad (36)$$

### 6.2.3 Optimisation results

Three cases were considered for optimisation using task 2 and were all run for 1000 optimisation iterations. The first optimisation case was to optimise the geometry of a standard corner with tight radii subject to a max thinning constraint of 0.1, and the results for this case are summarised in Figure 31. The performance history during the optimisation process is plotted in Figure 31(a). The blue curve shows that the initial design resulted in a max thinning strain of 0.45. Through modifying the geometry this value was quickly decreased to below 0.1 at iteration 56, i.e, to satisfy the imposed manufacturing constraint. The yellow curve shows that the similarity loss simultaneously increased up to iteration 56 as the max thinning decreased. This increase was due to the geometry becoming increasingly more dissimilar to the initial geometry to satisfy the constraint.

After iteration 56 in Figure 31(a), the similarly loss then progressed to decrease while the max thinning remained at the max thinning constraint value of 0.1. The mechanism for why max thinning curve remained at 0.1 was investigated. The zoomed in plot in Figure 31(a) reveals very minor oscillations below the max thinning constraint from iteration 56 onwards. To find the origin of these oscillations, the histories of the individual terms in the bracket of Equation (36), which drove the optimiser, were plotted, and are presented in Figure 31(b). It was found that the backward gradient term imposed large penalties on the total Lagrangian function each time the max thinning, shown in Figure 31(a), went above the max thinning constraint. Conversely, each time the max thinning fell below the constraint, the penalty was removed. After the penalty was removed, the max thinning began to increase again, and this was because the constraint was inactive, and the similarity loss was simultaneously decreasing. This periodic behaviour was due to the formulation in Equation (32) and restricted the optimiser from finding solutions with max thinning above the imposed constraint level.

Reconstructed geometries at three noteworthy stages of the optimisation are presented in Figure 31(c), and sections of these geometries are given in Figure 31(d & e) for further comparison. The geometries at iteration 56 and 1000 both satisfied 0.1 max thinning, but the latter was as close as possible to the initial geometry from iteration 0.

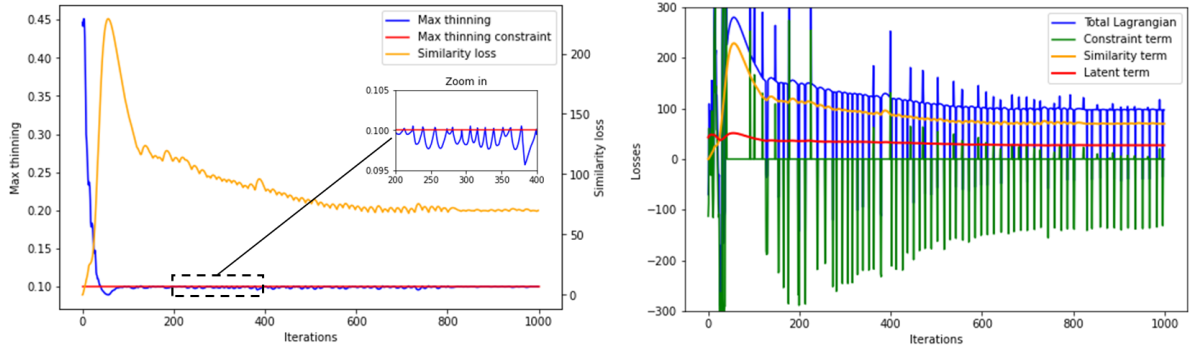

(a) Design performance vs optimisation iterations  (b) Lagrangian terms vs optimisation iterations

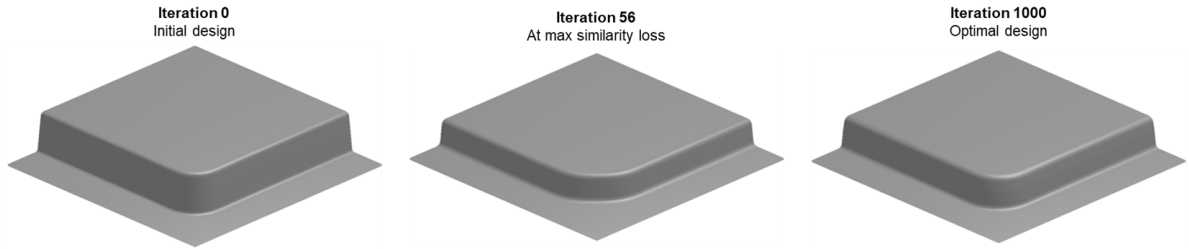

(c) Geometries before, during and after optimisation

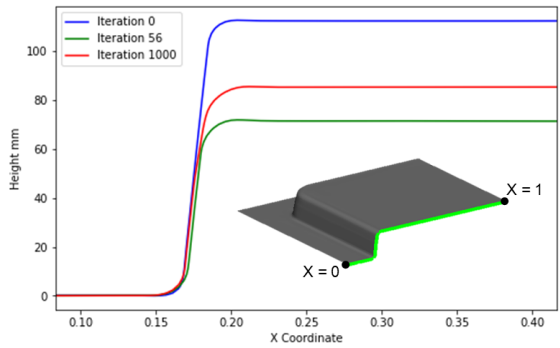 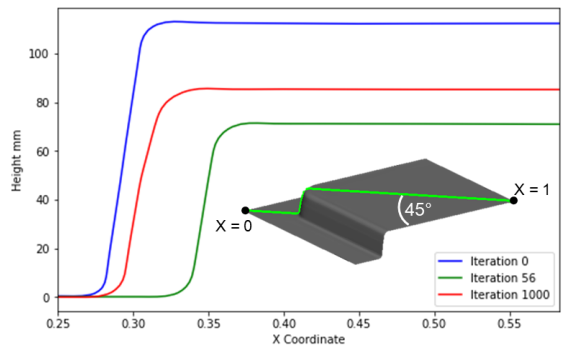

(d) 0° Section of geometries in (c)  (e) 45° Section of geometries in (c)

Figure 31 Results for optimising a standard corner geometry using the optimisation platform. Max thinning constraint set at 0.1.

The second optimisation case was to optimise a stepped sidewall geometry with tight radii subject to a max thinning constraint of 0.15 and the results for this case are summarised in Figure 32. Comparing the results in Figure 32 with Figure 31 demonstrates that the proposed optimisation platform is able to optimise geometries irrespective of CAD parameterisation scheme.

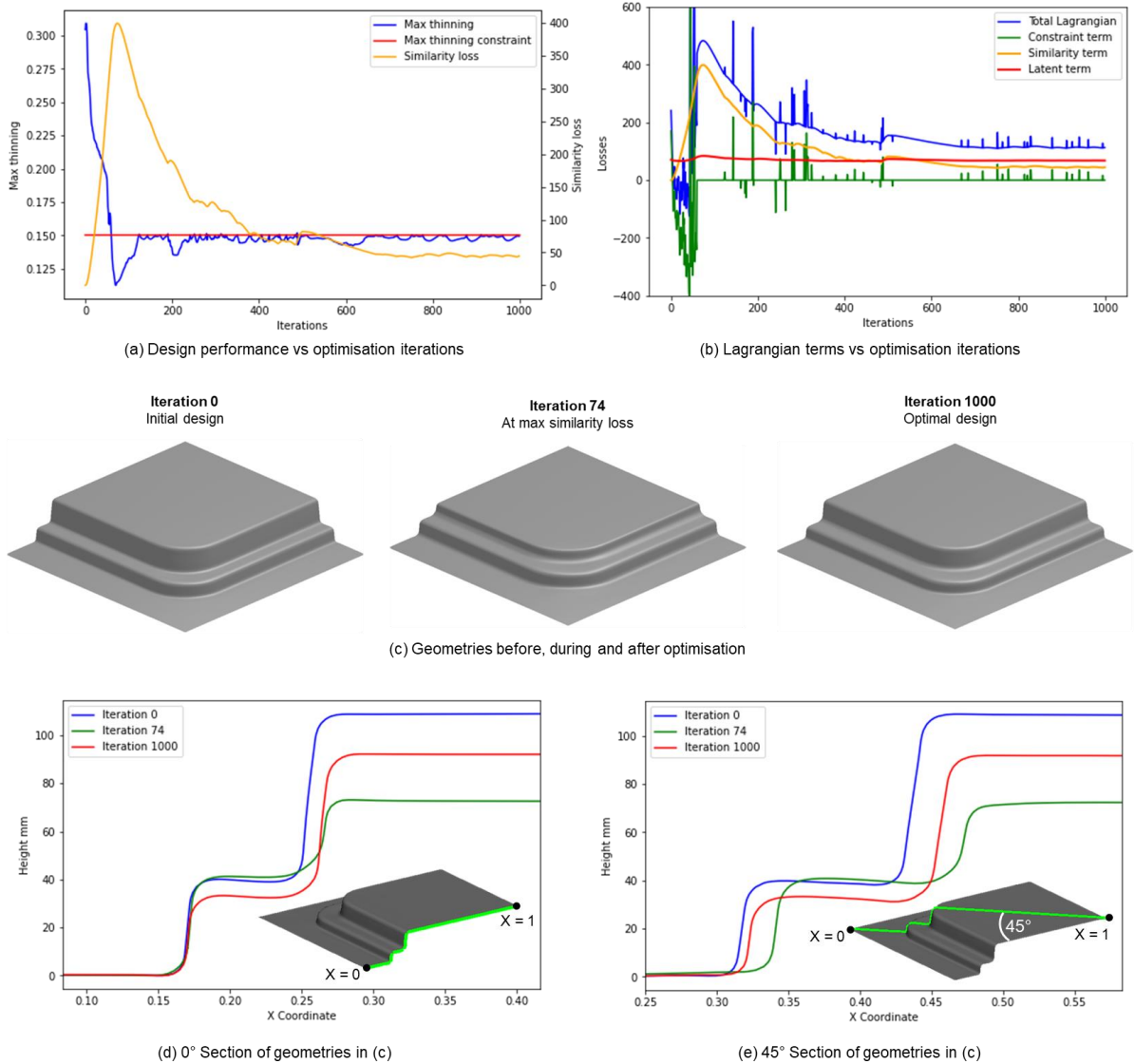

Figure 32 Results for optimising a stepped sidewall geometry using the optimisation platform. Max thinning constraint set at 0.15.

The third optimisation case was to optimise another stepped sidewall geometry with tight radii, but this time to satisfy a stricter max thinning constraint of 0.05 and the results for this case are summarised in Figure 33. Figure 33(a) again shows the simultaneous decrease in max thinning and increase in similarity loss up to iteration 174. However, a bend was found in the similarity loss curve at iteration 110 and is highlighted in the zoomed in subfigure. The max thinning at iteration 110 was 0.1, which was still above the 0.05 constraint. To meet this constraint, the optimiser had to change the geometry subclass from a stepped sidewall into a standard corner, and this resulted in the rapid increase in similarity loss from iteration 110 to 174. This change in geometry subclass is evidenced in Figure 33(c) when comparing the initial geometry at iteration 0 with the newly created standard corner at iteration 174. As the optimisation progressed further beyond iteration 174, the similarity loss progressed to decrease while the max thinning remained at the constraint level for the same reasons detailed earlier for Figure 31. The resulting geometry at iteration 1000 was like a standard corner with a new profile which is seen in Figure 33(d & e). Overall, the results summarised in Figure 33 demonstrate that optimisation of stamping geometries between geometric parameterisation schema is possible for the first time. This result means that the geometry is able to deform with complete freedom and that all constraints relating to CAD parameterisation scheme are removed. This removal results in increased design guidance in an industrial setting and potential discovery of new designs which are suitable for a selected stamping process.

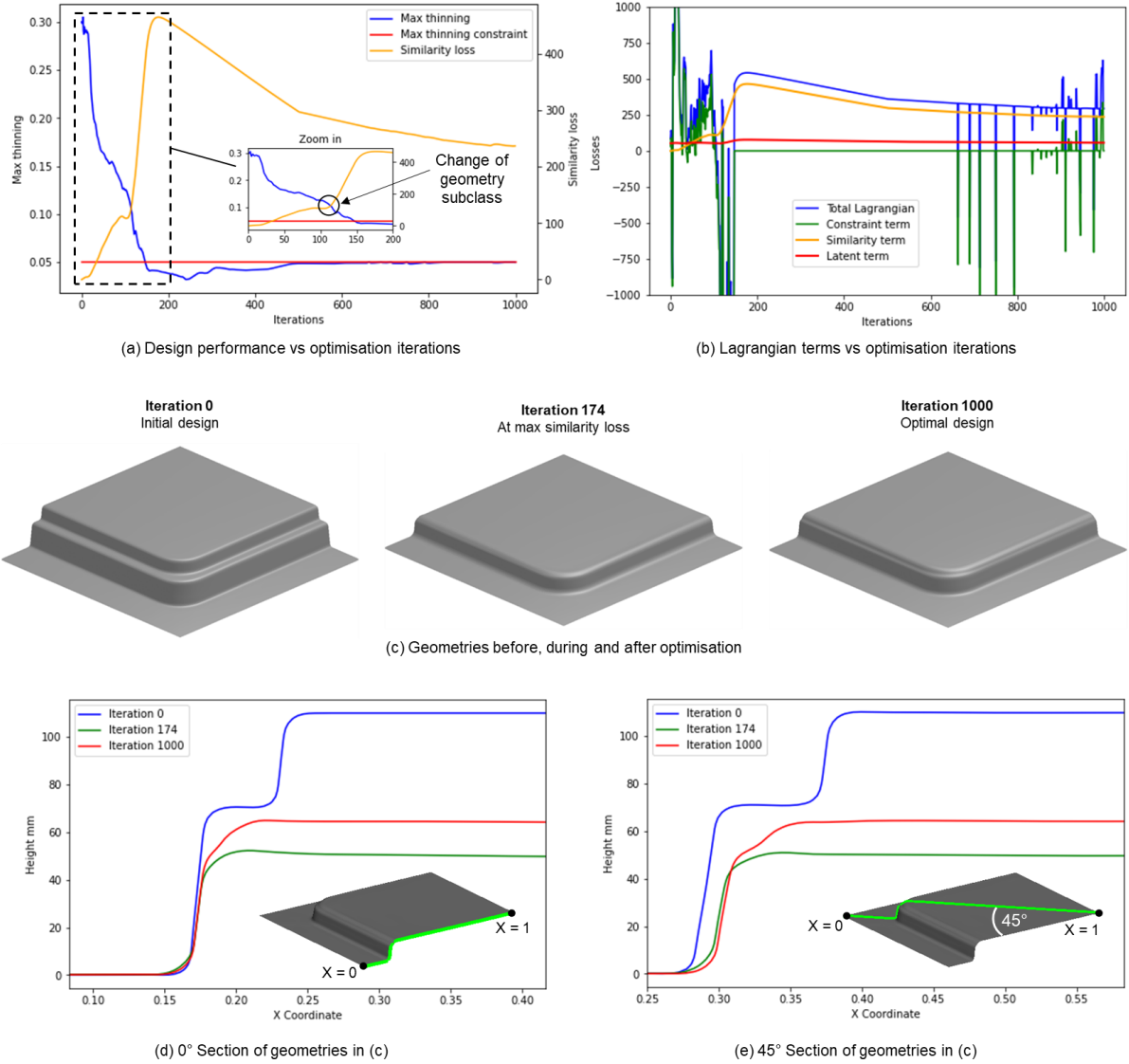

Figure 33 Results for optimising a stepped sidewall geometry using the optimisation platform. Max thinning constraint set at 0.05.

# 7    Conclusion and broader impact

This paper presented a comprehensive study on the development of a new deep-learning based platform for optimising sheet stamping geometries subject to manufacturing constraints. Based on the conducted research, the following conclusions can be summarised:

- A new computer based method for modelling sheet stamping geometries using signed distance field (SDF) based implicit neural representations was established. The challenges of using implicit neural representations for these geometries was discussed and addressed through novel network training strategies which exploited geometric properties of SDFs. These strategies enabled high reconstruction accuracy of local geometric features, such as tight radii, which play a critical role in sheet stamping performance. The high reconstruction accuracy using the new approach was verified through a comprehensive quantitative and qualitative evaluation.

- Training and evaluation of a geometry generator network $f_{\theta_1}$ was performed using the newly developed method. This network was a multi-layer perceptron that modelled sheet stamping geometries using implicit neural representations for the first time. $f_{\theta_1}$ mapped a latent vector $z$ to a continuous SDF. The zero-level-set of the SDF implicitly represented the sheet stamping geometry that was encoded in $z$. Changing the latent vector (i.e., in an optimisation setting) changed the generated SDF, and this phenomenon was exploited for the optimisation of sheet stamping geometries.

- Training and evaluation of a manufacturing performance evaluator network $g_{\theta_2}$ was performed. This network was a CNN based surrogate model of a hot stamping process. The model mapped a 2D projected height map of the geometry reconstructed from $f_{\theta_1}$ to its post-stamping thinning field. The thinning field was used as a manufacturing performance indicator for driving the optimisation of sheet stamping geometries.

- The optimisation platform was constructed from the interaction of $f_{\theta_1}$ and $g_{\theta_2}$. The platform featured a forward and backward pass through both networks. The forward pass enabled an optimisation task-specific objective function to be computed. The backward pass enabled the manufacturing performance information from $g_{\theta_2}$ to be used to iteratively update the geometries generated by $f_{\theta_1}$ using a gradient based optimisation technique. These updates were made to minimise the objective function and thus improve manufacturing performance.

- The optimisation platform was used to optimise deep drawn corner geometries subject to a manufacturing constraint on maximum post-stamped thinning strain from a hot stamping process. A new objective function was formulated which minimised the difference between the optimum and the initial geometries while ensuring the manufacturing constraint was met. The results showed that expressive geometry changes between geometry subclasses were achievable for the first time. These changes were agnostic to geometric complexity and were driven by stamping performance.

The proposed optimisation platform has the potential to have a major impact in sheet stamping industries. Non-trivial geometric changes to existing components to suit new stamping processes can be brought to light. In addition, complex geometries with hundreds of CAD dimensions and built from different geometric parameterisation schema, such as door inners or B-Pillars, could be optimised quickly and effectively.

## Acknowledgements

The authors thank the funding support by Impression Technologies Ltd and the UK Engineering and Physical Sciences Research Council. Software and technical support from Rajab Said and Mustapha Ziane from ESI Group is also gratefully acknowledged. HFQ® is a registered trademark of Impression Technologies Ltd.

## Appendix A: Brief explanation of Marching Cubes

It was seen that signed distance fields (SDFs) can be effective at *implicitly* modelling arbitrary shape boundaries and surfaces as their zero-level-sets. However, a number stamping applications demand *explicit* surface representations, such as meshes for CAD modelling or FE stamping simulations [6], and images as inputs to image-based surrogate models [31]. In this work, explicit representations in the form of 2D height map images of stamping die geometry meshes were needed to assess the manufacturing performance of these geometries by $g_{\theta_2}$, as shown in Figure 5.

Combining the benefits of SDFs with explicit representations requires the use of Marching Cubes [51]. Marching Cubes is an algorithm for creating a triangle mesh based representation of a level-set from an implicit field. Since the zero-level-set of SDFs here are designed to implicitly represent the surface geometry, running Marching Cubes on the SDF would extract the implicit surface by converting it to an explicit triangular mesh. This application of Marching Cubes works by iterating over a uniform 3D grid of cubes imposed over a region of the SDF and searching for its zero-level-set. For each cube in the 3D grid, the algorithm evaluates the sign of the SDF points at the cube vertices. If all 8 cube vertices of the cube have the same sign, then the cube is entirely above or below the zero-level-set and no triangles are created. Otherwise, triangles are created which make up mesh faces. There are 15 unique cases of triangle and mesh vertex combinations and these are determined by a lookup table shown in Figure A.1, where a search is performed using the SDF sign at each of the 8 cube vertices. The final extracted mesh is a combination of the triangles and mesh vertices from each cube in the 3D grid. For more details on Marching Cubes, the reader is referred to the original paper [51].

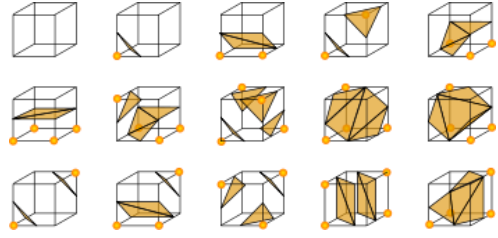

Figure A.1 A graphical representation of the triangulated cubes lookup table used by the Marching Cubes algorithm. Yellow highlighted cube vertices have an opposite to non-highlighted cube vertices. There are 15 unique combinations in total.

## Appendix B: Variants of Auto-Decoders for learning SDFs

Three different Auto-Decoder architectures were considered in this study, and these are shown in Figure A.2. These architectures were inspired by the result presented by Park *et al.* in that increasing the number of skip connections and network depth improved regression accuracy on their dataset [48]. A description Auto-Decoder A is provided in Section 4.1, and Auto-Decoders B and C follow similar descriptions.

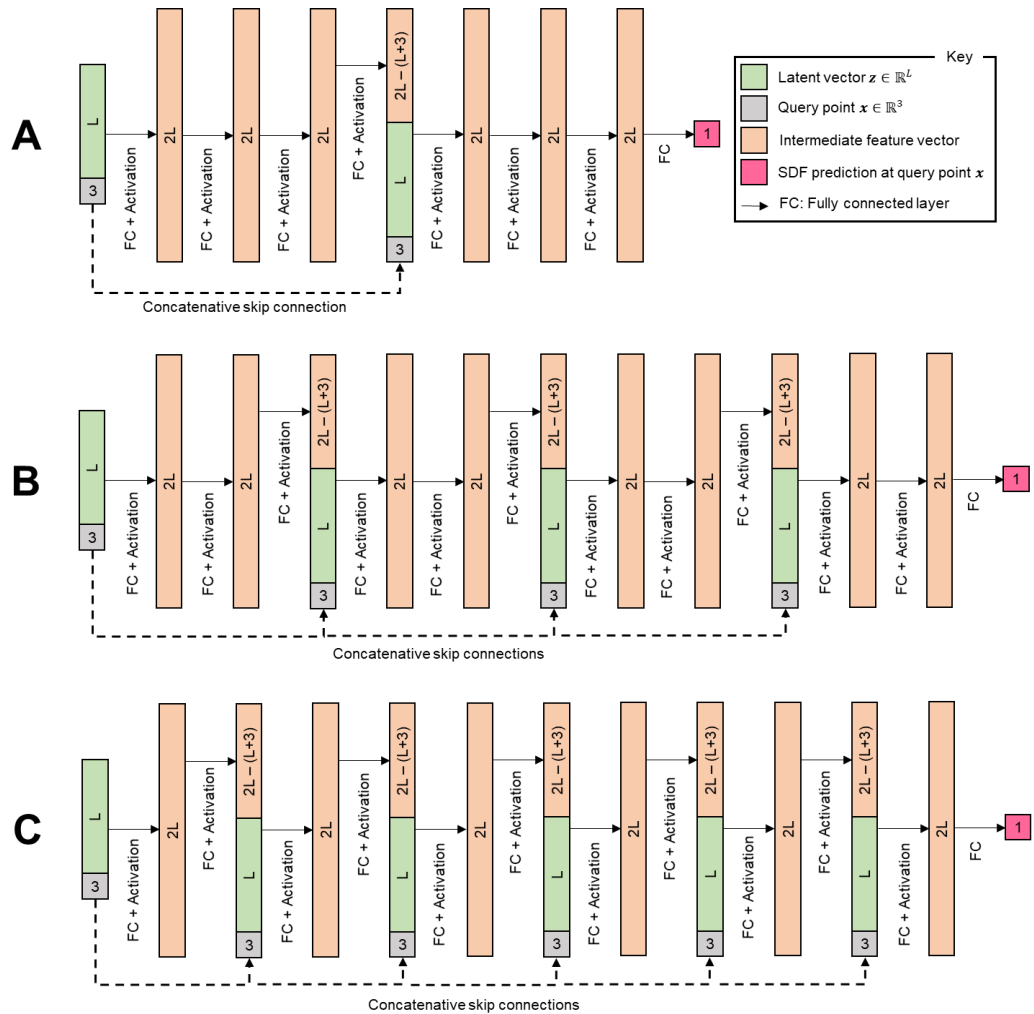

Figure A.2 Three Auto-Decoder architectures which were used to generate an SDF value from an input 3D point and a conditioning latent vector. The numbers denote vector lengths.

Based on these three architectures five different variants of Auto-Decoders were trained and their performances compared. These variants are summarised in Table A.1.

Table A.1 Variants of Auto-Decoders used in this study, based on the architectures from Figure A.2 and training approaches introduced in Section 4.2.

| Network name   | Architecture | Latent vector length $L$ | Activation function | Training approach |
|----------------|--------------|--------------------------|---------------------|-------------------|
| Explicit Net 1 | A            | 128                      | ReLU                | Explicit          |
| Explicit Net 2 | A            | 32                       | ReLU                | Explicit          |
| Explicit Net 3 | B            | 128                      | ReLU                | Explicit          |
| Explicit Net 4 | C            | 128                      | ReLU                | Explicit          |
| Implicit Net   | A            | 128                      | SoftPlus            | Implicit          |